\documentclass{article}

 \usepackage[main, final]{neurips_2026}

\usepackage[utf8]{inputenc} 
\usepackage[T1]{fontenc}    
\usepackage{hyperref}       
\usepackage{url}            
\usepackage{booktabs}       
\usepackage{amsfonts}       
\usepackage{enumitem}
\usepackage{nicefrac}       
\usepackage{microtype}      
\usepackage{xcolor, colortbl}         
\usepackage{float}
\usepackage{multirow}
\usepackage{graphicx}
\usepackage{subfigure}
\usepackage{amsmath}
\title{GAD in the Wild: Benchmarking Graph Anomaly Detection under Realistic Deployment Challenges}
\author{Jingjing Zhou$^{1}$, Shiyu Huang$^{1}$, Qing Qing$^{2}$, Zuquan Yuan$^{2}$, Huafei Huang$^{3}$, Ziqi Xu$^{4}$,\\
\textbf{Mingliang Hou}$^{5}$, \textbf{Xikun Zhang}$^{4}$, \textbf{Renqiang Luo}$^{4}$, \textbf{Ivan Lee}$^{3}$\\
$^1$Zhejiang Gongshang University, $^2$Jilin University, $^3$Adelaide University, \\
$^4$RMIT University, $^5$ Jinan University
}

\begin{document}

\maketitle

\begin{abstract}
Graph Anomaly Detection (GAD) is a critical task in graph machine learning with vital applications in financial fraud detection and social platform governance. 
However, existing GAD benchmarks are often restricted to small-scale, curated graphs with relatively balanced anomaly ratios, leaving a substantial gap between academic evaluation and real-world deployment. 
To bridge this gap, we present a multi-dimensional benchmark that systematically evaluates GAD models under three deployment-relevant challenges: million-scale graphs, extreme anomaly scarcity, and missing node attributes.
We derive a family of controlled benchmark variants from five diverse graphs, including two native industrial-scale datasets with over $3.7$ million nodes. 
Our extensive evaluation of nine representative GAD models reveals three major limitations: (1) most GNN-based methods fail to scale to million-node graphs due to prohibitive memory requirements; (2) detection performance drops sharply under realistic anomaly ratios (e.g., $0.1\%$), often resulting in zero recall; and (3) reconstruction-based models are highly sensitive to attribute imputation strategies. 
Our findings suggest that strong performance in laboratory settings does not guarantee robustness in production environments. 
We release this benchmark and empirical evaluation as a diagnostic testbed to promote the development of robust and scalable GAD systems for large-scale, imperfect graphs encountered in practice.
Code is available at~\url{https://anonymous.4open.science/r/Benchmark_GAD-E7A3}. 
\end{abstract}

\section{Introduction}
\par Graph Anomaly Detection (GAD)~\citep{QiaoTAKAP25,RenXLHA23} aims to identify nodes or substructures that deviate significantly from normal patterns in graph-structured data. 
It is a fundamental task in machine learning with vital applications in financial fraud detection~\citep{huang2022dgraph,ChenT22}, misinformation tracking~\citep{Dou22}, and social platform governance~\citep{WangZ22}. 
In these scenarios, anomalies are rare but carry high risk.
Therefore, building accurate GAD models is essential for maintaining trustworthy systems and effective risk control~\citep{ZhaoSWZWWL26}. 
As real-world systems grow in scale and complexity, GAD methods must not only achieve strong detection performance but also remain reliable under realistic deployment conditions~\citep{CaoXZWYW025}. 
This makes representative and practically grounded benchmarks indispensable to progress in the field.

\par Despite the rapid development of GAD methods, current research remains heavily centered on a few legacy datasets~\citep{ding2019dominant,YuanZYHC021}. 
These datasets often fail to reflect the reality of modern data, which is increasingly large, noisy, and incomplete. 
In particular, commonly used datasets are typically small in scale (often fewer than $100$k nodes), have relatively high anomaly ratios (on average above $3$\%), and assume complete data without missing values, all of which differ from real-world scenarios.
While traditional benchmarks offer controlled environments for comparison, they often feature idealized settings that do not exist in production~\citep{LiuDZDHZDCPSSLC22,gui2022good}. 
For instance, many widely used datasets exhibit unrealistically high anomaly ratios and perfectly clean node attributes. 
Consequently, models that excel in these "laboratory" settings may fail when deployed in noisy, real-world environments~\citep{neo2024fairgad}.
This mismatch exposes a persistent gap between existing academic evaluations and the requirements of real-world applications~\citep{BoumanBH24}.

\par Furthermore, the problem of "low-quality" data, such as missing features or extreme class imbalance, remains largely unexplored in existing benchmarks. 
A model might scale well to millions of nodes but collapse when $50$\% of the attributes are missing or when the anomaly rate drops to $0.1$\%. 
Without evaluating these factors jointly, we cannot understand the true boundaries of current GAD algorithms.

\par To bridge this gap, we construct three types of datasets that simulate the challenges of real-world data: large-scale graphs, missing node attributes, and extreme anomaly rarity.. 
Starting from five diverse graph datasets, we derive controlled benchmark variants by increasing graph scale, introducing attribute missingness, and adjusting anomaly prevalence to better reflect real-world conditions.
Specifically, we simulate three critical scenarios: (1) Large-scale graphs to test hardware efficiency; (2) Incomplete data with missing node attributes to test structural resilience; and (3) Extreme imbalance to see if models can still identify very rare anomalies. 
This multi-dimensional approach provides a more rigorous and realistic stress test for GAD models than traditional benchmarks.

\par We conduct extensive experiments using a wide range of GAD methods on these new datasets. 
Our analysis focuses on three key metrics: scalability under medium-scale compute, performance after basic attribute imputation, and the raw number of anomalies recovered under low-ratio settings. 
Unlike previous work that prioritizes top-line accuracy, we explicitly examine memory consumption and robustness to imperfect data. 
Our results uncover important limitations of current methods and provide practical insights for deploying GAD models in real-world, resource-constrained environments.
In summary, our contributions are threefold:
\begin{itemize} [leftmargin=0.5cm]
    \item We formalize a comprehensive realism gap in GAD evaluation that extends beyond mere graph size. 
    We argue that realistic assessment should simultaneously account for computational scalability, extreme class imbalance, and attribute incompleteness.
    \item We develop a multi-dimensional benchmark featuring controlled dataset variants and native industrial-scale graphs. 
    This framework enables a rigorous, isolated evaluation of models under diverse, deployment-oriented constraints.
    \item Comprehensive Empirical Insights: We provide a unified study of representative GAD models, assessing both detection efficacy and hardware efficiency. 
    Our findings reveal critical robustness trade-offs and practical limitations that were previously overlooked in conventional, idealized benchmark settings.
\end{itemize}

\section{Data Construction and Processing}
\par This section introduces the data preprocessing, variant, and quality-control procedures used to construct our multi-dimensional GAD benchmark.
We utilize five public GAD datasets spanning social media, academic citations, and financial fraud. 
These are categorized into three foundational datasets for controlled variants (Twitter, PubMed, and Credit) and two industrial-scale graphs (DGraph-Fin and T-Social) specifically for scalability testing. 
All datasets undergo a unified preprocessing pipeline that ensures topological integrity (removing duplicate edges/self-loops), feature standardization (min-max normalization to $32$-bit floats), and label consistency. 
The original graphs provide the reference statistics for all subsequent controlled experiments; detailed dataset descriptions and the summary in Table~\ref{tab:basic_stats} are provided in Appendix~\ref{sec:data}.

\par We focus on the three realism gaps identified in the introduction: limited graph scale, unrealistic anomaly ratios, and idealized complete attributes.
To bridge the realism gaps identified in the introduction, we construct three specialized experimental scenarios:
Scale-expanded Benchmarking: We extend the evaluation breadth by incorporating three foundational datasets alongside two industrial-scale graphs. 
This hierarchy allows us to evaluate model performance across both controlled laboratory settings and massive-scale real-world environments.
Anomaly Ratio Adjustment: For the three small-to-medium datasets, we recalibrate the anomaly labels to more realistic, sparse distributions, specifically controlling the anomaly ratio to below $1\%$. 
This adjustment represents the nature of real-world GAD tasks where anomalies are rare.
Attribute Incompleteness and Imputation: To simulate real-world data quality issues, we introduce controlled levels of feature corruption and missing values into the three foundational datasets. 
This allows us to construct diverse scenarios for testing the robustness of various imputation strategies.

\par Notably, the latter two scenarios (Ratio Adjustment and Attribute Incompleteness) are primarily executed on the three foundational datasets. 
This strategic choice is necessitated by hardware constraints. 
Industrial datasets provide critical scalability insights, but they frequently encounter out-of-memory (OOM) errors under standard laboratory computational budgets. 
This constraint aligns with the hardware configurations reported in most current baselines.
By focusing sensitivity analyses on foundational graphs, we ensure a comprehensive and reproducible evaluation of algorithmic behavior without the confounding factor of hardware-induced failures.

\begin{table}[t]
    \centering
    \renewcommand{\arraystretch}{1.25}
    \caption{Statistics of Basic Graph Anomaly Detection Datasets}
    \footnotesize
    \begin{tabular}{lrrrrl}
      \toprule
      \textbf{Dataset} & \textbf{Nodes} & \textbf{Edges} & \textbf{Attribute Dim.} & \textbf{Anomaly Ratio} & \textbf{Task Type} \\
      \midrule
      Twitter   & $47,712$   & $468,697$   & $780$ & $6.66\%$  & Misinformation Detection \\
      PubMed    & $19,717$   & $44,338$    & $500$ & $3.04\%$  & Abnormal Paper Detection \\
      Credit    & $30,000$   & $1,436,858$ & $23$  & $22.12\%$ & Credit Default Detection \\
      DGraph-Fin & $3,700,550$ & $4,300,999$ & $17$ & $1.27\%$  & Financial Fraud Detection \\
      T-Social   & $5,781,065$ & $73,105,508$ & $10$ & $3.01\%$  & Abnormal Social User Detection \\
      \bottomrule
    \end{tabular}
    \vspace{-2em}
  \label{tab:basic_stats}
\end{table}

\subsection{Scale-Expanded Datasets}
\par Most GAD benchmarks rely on small graphs, which fail to reflect the challenges of large-scale deployment. 
To bridge this gap, we construct scale-expanded datasets that increase the number of nodes and edges while strictly preserving the original data distribution.

\par Our expansion pipeline consists of three main steps. 
First, we analyze the foundational datasets to characterize their original anomaly ratios, degree distributions, and attribute patterns. 
Second, we generate new normal and anomalous nodes to reach target sizes of $500$k and $1$M nodes. 
Their attributes are sampled from Gaussian Mixture Models (GMMs) fitted to the original data. 
Third, we use a hierarchical edge construction strategy to preserve connection patterns. 
We stratify nodes into core, middle, and edge layers based on the original degree distribution and add edges between layers to maintain the original edge-to-node ratio.

\par Furthermore, we include two native industrial datasets, DGraph-Fin and T-Social, as complementary benchmarks. 
While our expanded datasets inherit the properties of smaller graphs, these native datasets provide a stress test on real-world, million-node structures, helping us evaluate deployment feasibility without synthetic artifacts.

\subsection{Anomaly Ratio Adjustment Datasets}
\par In real-world scenarios, anomalies are far rarer than those found in standard benchmarks. 
High anomaly ratios simplify the detection task and can give an overly optimistic view of model performance under extreme imbalance.
We therefore construct ratio-adjusted datasets that modify only the anomaly proportion while keeping the topology and attribute distribution intact.

\par To reduce the anomaly ratio, we select which anomalous nodes to retain using three strategies:
\begin{itemize} [leftmargin=0.5cm]
    \item Core Cluster Retention: Keeps anomalies from the largest cluster in the attribute space to represent concentrated patterns.
    \item Edge Cluster Retention: Keeps anomalies from the smallest cluster to simulate scattered, rare cases.
    \item Random Retention: Selects anomalies uniformly at random.
\end{itemize}

\par For nodes that are converted from anomalous to normal, we update their labels and replace their attributes with the mean vector of their normal neighbors (or the global normal mean). 
This ensures that the only manipulated variable is the anomaly ratio. 
For each foundational dataset, we generate versions with anomaly ratios of $0.5\%$ and $0.1\%$ across all three strategies.

\subsection{Attribute Incompleteness and Imputation Datasets}
\par Standard benchmarks typically assume complete node attributes, yet missing data is pervasive in practice~\citep{LuoHTRXHDX26}. 
Since most GAD models cannot handle missing values directly, they rely on imputation. 
We simulate this by constructing datasets with controlled missing ratios and multiple imputation strategies.

\par We inject missingness by applying separate masks to normal and anomalous nodes, allowing us to simulate category-dependent data gaps. 
To ensure the data is compatible with mainstream GAD models, we apply three common imputation methods:
\begin{itemize} [leftmargin=0.5cm]
    \item Mean Imputation~\citep{you2020grape}: Fills gaps with the global category mean. 
    \item Median Imputation~\citep{rossi2022featureprop}: Fills gaps with the global category median.
    \item Neighbor Imputation~\citep{um2023pcfi}: Fills gaps with the mean attribute of neighbors from the same category.
\end{itemize}

\par We generate variants with missing ratios ranging from $10\%$ to $50\%$. 
By combining these ratios with different imputation strategies, we can systematically study how GAD models react to both the degree of missing data and the choice of preprocessing method.

\section{Methodology}
\par We evaluate GAD models under three deployment-oriented challenges: \textbf{scalability}, \textbf{imbalance robustness}, and \textbf{incomplete-data robustness}.
Our evaluation follows a strict \textit{variable-isolation} principle.
For each task, we adjust only the target dimension while keeping all other data distributions and experimental settings fixed.
This design ensures that observed performance differences can be attributed to the specific capability being tested.

\par Concretely, our benchmark is organized around three research questions:
\begin{itemize}  [leftmargin=0.5cm]
    \item \textbf{RQ1}: Can GAD models maintain stable detection performance and acceptable computational overhead when scaling to industrial large-scale graphs?
    \item \textbf{RQ2}: Can GAD models effectively detect rare anomalies under extreme class imbalance?
    \item \textbf{RQ3}: Can GAD models perform anomaly reasoning based on topological structure when attributes are incomplete, or do they over-rely on complete attribute settings?
\end{itemize}

\par All evaluations are repeated across transformed versions of the three foundational datasets, namely Twitter, PubMed, and Credit, to avoid overfitting conclusions to a single graph.
We follow the official default data split of each model implementation and partition nodes into non-overlapping training, validation, and test sets, denoted by $\mathcal{V}_{\text{train}}$, $\mathcal{V}_{\text{val}}$, and $\mathcal{V}_{\text{test}}$.
The validation set is used for early stopping, and all final metrics are reported on the held-out test set.

\par Formally, we denote sets with calligraphic uppercase letters (e.g., $\mathcal{A}$), matrices with bold uppercase letters (e.g., $\mathbf{A}$), vectors with bold lowercase letters (e.g., $\mathbf{a}$), and the cardinality of a set as $|\mathcal{A}|$.
Given an attributed graph $\mathcal{G}=(\mathcal{V},\mathcal{E},\mathbf{X})$, where $\mathcal{V}$ and $\mathcal{E}$ represent the sets of nodes and edges, respectively, $\mathcal{N}(v_i)$ denotes the set of neighbors for node $v_i$, and $\mathbf{X} \in \mathbb{R}^{|\mathcal{V}|\times d}$ is the node attribute matrix.
We follow indexing conventions similar to NumPy in Python:
$\mathbf{A}[i,j]$ denotes the element at the $i$-th row and the $j$-th column, while $\mathbf{A}[i,:]$ and $\mathbf{A}[:,j]$ represent the $i$-th row and the $j$-th column, respectively.
Each node $v_i \in \mathcal{V}$ is associated with a binary label $y_i \in \{0,1\}$, where $y_i = 0$ indicates a normal node and $y_i=1$ indicates an anomalous node.
Correspondingly, we denote the sets of normal nodes and anomalous nodes as $\mathcal{V}_0$ and $\mathcal{V}_1$, with their respective attribute matrices denoted as $\mathbf{X}_0$ and $\mathbf{X}_1$.

\par The goal of unsupervised GAD is to learn an anomaly scoring function $f_{\theta}:\mathcal{V}\rightarrow\mathbb{R}$, where a larger score indicates a higher anomaly likelihood.
Following the standard unsupervised setting adopted by representative GAD methods \citep{ding2019dominant,fan2020anomalydae,liu2022cola,roy2024gadnr}, models are trained on $\mathcal{V}_{\text{train}}$ without label supervision, tuned on $\mathcal{V}_{\text{val}}$ for early stopping, and evaluated on $\mathcal{V}_{\text{test}}$:
\begin{equation}
\theta^{*}=\arg\min_{\theta}\mathcal{L}\left(\left\{f_{\theta}(v)\mid v\in \mathcal{V}_{\text{train}}\right\}\right),
\end{equation}
where $\mathcal{V}_{\text{train}} \cap \mathcal{V}_{\text{val}} \cap \mathcal{V}_{\text{test}}=\emptyset$ and $\mathcal{V}_{\text{train}} \cup \mathcal{V}_{\text{val}} \cup \mathcal{V}_{\text{test}}=\mathcal{V}$.

\subsection{Scalability Evaluation Task}
\par For scalability evaluation, let $\mathcal{G}=(\mathcal{V},\mathcal{E},\mathbf{X})$ denote the original graph and let $\mathcal{G}^{s}=(\mathcal{V}^{s},\mathcal{E}^{s},\mathbf{X}^{s})$ denote a scale-expanded graph in our expanded benchmark suite.
We ensure that all variants differ only in node scale, while anomaly ratio, degree distribution, and attribute distribution remain consistent with the original graph:
\begin{equation}
    \begin{aligned}
        & \frac{|\mathcal{V}^s_1|}{|\mathcal{V}^s|} = \frac{|\mathcal{V}_1|}{|\mathcal{V}|}, \\
        p_{\text{KS}}\bigl(\mathrm{deg}(\mathcal{G}^{s}),\mathrm{deg}(\mathcal{G})\bigr) > 0.05, \quad 
        &p_{\text{KS}}\bigl(\mathbf{X}^{s}_{0},\mathbf{X}_{0}\bigr) > 0.05,
        \quad
        p_{\text{KS}}\bigl(\mathbf{X}^{s}_{1},\mathbf{X}_{1}\bigr) > 0.05,
    \end{aligned}
\end{equation}
where $\mathrm{deg}(\mathcal{G})$ denotes the degree distribution of graph $\mathcal{G}$, and $p_{\text{KS}}$ is the p-value of the Kolmogorov-Smirnov (KS) test (ensuring no significant distribution shift).

\par We evaluate scalability from two perspectives.
First, performance stability measures detection performance on the original graph and its $500$k- and $1000$k-node variants using AUC-ROC and AUC-PR.
Second, computational efficiency records end-to-end training time and peak GPU memory usage.
In addition, we repeat the same evaluation pipeline on the two native industrial-scale datasets, DGraph-Fin and T-Social, to assess real deployment feasibility beyond controlled graph expansion.

\subsection{Imbalance Robustness Evaluation Task}
\par For imbalance robustness, let $\mathcal{G}^r$ denote a ratio-adjusted graph, where $\mathcal{R}$ denotes the retained anomaly set.
All graph variants preserve topology and overall attribute structure, and differ only in anomaly ratio and anomaly distribution pattern.
For selected anomaly nodes, we perform label and attribute conversion as follows:
\begin{equation}
    \begin{aligned}
    y_i^r &=
    \begin{cases}
    1, & v_i \in S, \\
    0, & \text{otherwise},
    \end{cases} \\
    \mathbf{X}^r[i,:] &=
    \begin{cases}
    \mathbf{X}[i,:], & v_i \in \mathcal{R} \\
    \mathbf{X}[i,:], & v_i \in \mathcal{G}_0, \\
    \mu({\mathcal{N}(v_i)\cap \mathcal{G}_0}), & v_i \in \mathcal{G}_1 \text{ and } v_i \notin \mathcal{R} \text{ and } \mathcal{N}(v_i)\cap \mathcal{G}_0 \neq \emptyset, \\
    \mu(\mathcal{G}_0), & \text{otherwise},
    \end{cases}
    \end{aligned}    
\end{equation}
$\mu(\cdot)$ denotes the mean function.

\par We evaluate imbalance robustness through two sub-tasks.
Anomaly Ratio Sensitivity measures performance under the original ratio, $0.5\%$, and $0.1\%$, with a particular focus on anomaly recall.
Anomaly Distribution Pattern Robustness compares performance under core-cluster, edge-cluster, and random anomaly-retention strategies to determine whether models can generalize beyond typical anomaly patterns.

\subsection{Incomplete-Data Robustness Evaluation Task}
\par For incomplete-data robustness, let $\mathcal{G}^m$ denote a graph with missing values.
We define an attribute-missing mask $\mathbf{M} \in \mathbb{R}^{|\mathcal{V}|\times d}$, where $\mathbf{M}[i,j]=1$ indicates that the $j$-th attribute of node $v_i$ is missing.
Correspondingly, the masks of normal and anomalous nodes are denoted as $\mathbf{M}_0$ and $\mathbf{M}_1$, respectively.
We generate category-specific missing masks according to:
\begin{equation}
    \begin{aligned}
        \gamma_0 = \frac{\sum\mathbf{M}_0[i,j]}{|\mathcal{V}_0| \times d}, \quad        
        \gamma_1 = \frac{\sum\mathbf{M}_1[i,j]}{|\mathcal{V}_1| \times d},
    \end{aligned}
\end{equation}
where $\gamma_0$ and $\gamma_1$ are the missing ratios for normal and anomalous nodes, respectively.
All graph variants preserve topology and anomaly ratio and differ only in the missingness pattern.

\par We evaluate incomplete-data robustness through two sub-tasks.
Missing Ratio Sensitivity measures model performance under missing ratios from $10\%$ to $50\%$.
Filling Strategy Adaptability compares mean, median, and neighbor imputation to determine whether model performance is overly sensitive to preprocessing choices.

\section{Experiments}
\subsection{Experimental Protocol and Baseline Models}
\par To ensure a rigorous and reproducible comparison, we establish a Unified Experimental Protocol. 
We fix the global random seed to $20$ for all data splitting and model initialization, reporting the average results over three independent runs. 
To avoid over-tuning bias, we strictly follow the Native Configuration Alignment: we adopt the official hyperparameter settings, data splits, and early-stopping strategies provided by the original authors of each model. 

\par All experiments are conducted on a standardized high-performance computing cluster to ensure that efficiency metrics, such as runtime and memory usage, are directly comparable.
Our hardware configuration is equivalent to that used by the baselines; specifically, all tasks are executed on an Intel(R) Xeon(R) Gold $6455$B CPU and an NVIDIA L$40$-$48$GB GPU.
The software stack consists of Ubuntu $22.04$, Python $3.8.20$, PyTorch $2.1.0$+cu$121$, and PyG (PyTorch Geometric) $2.6.1$.
For further details regarding the hardware specifications of all baselins, please refer to Appendix~\ref{sec:hardware}.

\par We assess models across two dimensions: detection effectiveness and deployment feasibility.
We report AUC-ROC, AUC-PR, and Anomaly Recall as effectiveness metrics. 
Specifically, Anomaly Recall is emphasized in our low-anomaly-ratio experiments to evaluate a model's ability to capture extremely rare instances.
To evaluate scalability under resource constraints, we record the end-to-end Training Time (seconds) and Peak GPU Memory Usage (GB) as efficiency metrics.
By reporting these metrics separately, we aim to provide a transparent view of the trade-offs between a model's predictive power and its practical computational cost.

\par We evaluate nine representative GAD models categorized into two groups:
Fundamental baselines: MLPAE (attribute-only autoencoder) and GCNAE~\citep{KipfW17} (standard GNN-based autoencoder).
State-of-the-art GAD models: We include dual-autoencoder structures (DOMINANT~\citep{ding2019dominant}, AnomalyDAE~\citep{fan2020anomalydae}), contrastive learning (CoLA~\citep{liu2022cola}), neighborhood reconstruction (GADNR~\citep{roy2024gadnr}), and high-performance methods targeting heterophily and scalability (AHFAN~\citep{wang2025ahfan}, SmoothGNN~\citep{dong2025smoothgnn}, and Huge-GAD~\citep{pan2025huge}). 
Detailed descriptions of each model are provided in Appendix~\ref{sec:baseline}.

\begin{table}[t]
    \centering
    \footnotesize
    \renewcommand{\arraystretch}{1.25}
    \begin{tabular}{lrrrrrrr}
    \toprule
    Datasets & Pubmed & Credit & Twitter & Twitter (Expand) & Twitter (Expand) & DGraph-Fin & T-Social \\
    \# nodes & $19$k & $30$k & $47$k & $500$k & $1,000$k & $3,700$k & $5,781$k \\
    \midrule
    MLPAE & \cellcolor{green!20} $6.05$ & \cellcolor{green!20} $3.08$ & \cellcolor{green!20} $5.00$ & \cellcolor{green!20} $52.60$ & \cellcolor{green!20} $110.33$ & \cellcolor{green!20} $369.97$ & \cellcolor{green!20} $633.42$ \\
    GCNAE & \cellcolor{green!20} $4.69$ & \cellcolor{green!20} $23.63$ & \cellcolor{green!20} $13.22$ & \cellcolor{green!20} $190.10$ & \cellcolor{red!20} OOM & \cellcolor{green!20} $494.67$ & \cellcolor{red!20} OOM \\
    COLA & \cellcolor{green!20} $3.49$ & \cellcolor{green!20} $25.45$ & \cellcolor{green!20} $9.08$ & \cellcolor{green!20} $127.61$ & \cellcolor{green!20} $270.90$ & \cellcolor{green!20} $568.83$ & \cellcolor{red!20} OOM \\
    GADNR & \cellcolor{green!20} $103.67$ & \cellcolor{green!20} $462.04$ & \cellcolor{green!20} $212.58$ & \cellcolor{red!20} OOM & \cellcolor{red!20} OOM & \cellcolor{red!20} OOM & \cellcolor{red!20} OOM \\
    DOMINANT & \cellcolor{green!20} $16.73$ & \cellcolor{green!20} $50.93$ & \cellcolor{green!20} $147.44$ & \cellcolor{red!20} OOM & \cellcolor{red!20} OOM & \cellcolor{red!20} OOM & \cellcolor{red!20} OOM \\
    AnomalyDAE & \cellcolor{green!20} $17.92$ & \cellcolor{green!20} $51.90$ & \cellcolor{red!20} OOM & \cellcolor{red!20} OOM & \cellcolor{red!20} OOM & \cellcolor{red!20} OOM & \cellcolor{red!20} OOM \\
    AHFAN & \cellcolor{green!20} $14.65$ & \cellcolor{green!20} $90.49$ & \cellcolor{green!20} $32.81$ & \cellcolor{green!20} $407.51$ & \cellcolor{green!20} $654.04$ & \cellcolor{green!20} $1913.98$ & \cellcolor{red!20} OOM \\
    SmoothGNN & \cellcolor{green!20}$3.69$ & \cellcolor{green!20} $4.21$ & \cellcolor{green!20} $9.01$ & \cellcolor{green!20} $54.41$ & \cellcolor{red!20} OOM & \cellcolor{green!20} $318.21$ & \cellcolor{red!20} OOM \\
    HUGE-GAD & \cellcolor{green!20}$63.36$ & \cellcolor{green!20} $203.17$ & \cellcolor{green!20} $78.86$ & \cellcolor{red!20} OOM & \cellcolor{red!20} OOM & \cellcolor{red!20} OOM & \cellcolor{red!20} OOM \\
    \bottomrule
    \end{tabular}
    \caption{Comparison of runtime performance (in seconds) across methods on multiple datasets.
    Green bars indicate that the model runs successfully on the current hardware; red bars indicate the model failed due to OOM errors.}
    \label{tab:runtime}
    \vspace{-2em}
\end{table}

\begin{table}[t]
    \centering
    \footnotesize
    \renewcommand{\arraystretch}{1.25}
    \begin{tabular}{lrrrrrrr}
    \toprule
    Datasets & Pubmed & Credit & Twitter & Twitter (Expand) & Twitter (Expand) & DGraph-Fin & T-Social \\
    \# nodes & $19$k & $30$k & $47$k & $500$k & $1,000$k & $3,700$k & $5,781$k \\
    \midrule
    MLPAE & \cellcolor{green!20} $330$ & \cellcolor{green!20} $226$ & \cellcolor{green!20} $970$ & \cellcolor{green!20} $8,626$ & \cellcolor{green!20} $17,212$ & \cellcolor{green!20} $9,074$ & \cellcolor{green!20} $18,312$ \\
    GCNAE & \cellcolor{green!20} $714$ & \cellcolor{green!20} $2,416$ & \cellcolor{green!20} $4,106$ & \cellcolor{green!20} $42,852$ & \cellcolor{red!20} OOM & \cellcolor{green!20} $11,666$ & \cellcolor{red!20} OOM \\
    COLA & \cellcolor{green!20} $320$ & \cellcolor{green!20} $2,374$ & \cellcolor{green!20} $1,032$ & \cellcolor{green!20} $10,644$ & \cellcolor{green!20} $21,254$ & \cellcolor{green!20} $16,478$ & \cellcolor{red!20} OOM \\
    GADNR & \cellcolor{green!20} $7,532$ & \cellcolor{green!20} $14,498$ & \cellcolor{green!20} $18,616$ & \cellcolor{red!20} OOM & \cellcolor{red!20} OOM & \cellcolor{red!20} OOM & \cellcolor{red!20} OOM \\
    DOMINANT & \cellcolor{green!20} $10,502$ & \cellcolor{green!20} $23,084$ & \cellcolor{green!20} $44,510$ & \cellcolor{red!20} OOM & \cellcolor{red!20} OOM & \cellcolor{red!20} OOM & \cellcolor{red!20} OOM \\
    AnomalyDAE & \cellcolor{green!20} $12,072$ & \cellcolor{green!20} $25,776$ & \cellcolor{red!20} OOM & \cellcolor{red!20} OOM & \cellcolor{red!20} OOM & \cellcolor{red!20} OOM & \cellcolor{red!20} OOM \\
    AHFAN & \cellcolor{green!20} $265$ & \cellcolor{green!20} $4,528$ & \cellcolor{green!20} $1,320$ & \cellcolor{green!20} $10,663$ & \cellcolor{green!20} $21,300$ & \cellcolor{green!20} $21,392$ & \cellcolor{red!20} OOM \\
    SmoothGNN & \cellcolor{green!20}$1,929$ & \cellcolor{green!20} $ 2,590$ & \cellcolor{green!20} $4,516$ & \cellcolor{green!20} $ 30,531$ & \cellcolor{red!20} OOM & \cellcolor{green!20} $25,693$ & \cellcolor{red!20} OOM \\
    HUGE-GAD & \cellcolor{green!20}$4,946$ & \cellcolor{green!20} $12,792$ & \cellcolor{green!20} $12,989$ & \cellcolor{red!20} OOM & \cellcolor{red!20} OOM & \cellcolor{red!20} OOM & \cellcolor{red!20} OOM \\
    \bottomrule
    \end{tabular}
    \caption{Comparison of memory usage (in MBs) across methods on multiple datasets.
    Green bars indicate that the model runs successfully on the current hardware; red bars indicate the model failed due to OOM errors.}
    \label{tab:memory}
    \vspace{-2em}
\end{table}

\subsection{Scalability Analysis: Most GAD Models Fail on Large-Scale Graphs}

\par Our scalability evaluation reveals a severe deployment bottleneck in current GAD research, as many models cannot handle industry-scale data.
When the graph scale expands to $500$k nodes, $4$ to $5$ out of $9$ models fail due to OOM errors across the controlled expanded datasets.
At $1,000$k nodes, the number of OOM models further increases to $4$ to $7$ out of $9$, depending on the dataset.
On the native T-Social dataset, only the attribute-only baseline MLPAE remains runnable, while all GNN-based models fail under the increased computational load.
These results suggest that most existing GAD methods are not yet ready for the million-node scale required by practical applications.
For comprehensive results, please refer to the Appendix~\ref{sec:scalability}, while specific training times and peak memory usage are summarized in Table~\ref{tab:runtime} and Table~\ref{tab:memory}.

\par Training efficiency is driven not only by the number of nodes but also by the complexity of node attributes.
Interestingly, the native DGraph-Fin dataset ($3,700$k nodes) occasionally consumes less GPU memory than our expanded Twitter dataset ($1,000$k nodes).
This occurs because GAD models must load and process all node features during training; the lower dimensionality of DGraph-Fin significantly reduces memory pressure compared to high-dimensional datasets.

\par For the few models that remain runnable, detection performance often becomes unstable as the graph scale grows.
The performance trend is dataset-dependent: some models degrade after scaling, while others improve on specific datasets such as PubMed.
For example, GCNAE suffers a $5.22$-point AUC-ROC drop on the Credit dataset when scaling from $30$k to $500$k nodes, while it improves substantially on the PubMed expanded graphs.
The attribute-only MLPAE remains relatively stable across scales, but its detection performance is generally limited.
These findings indicate that topology-aware models are vulnerable to scalability constraints and may become unreliable in realistic, large-scale scenarios.

\subsection{Imbalance Analysis: Current Models Struggle with Rare Anomalies}

\par
\begin{figure*}[t]
	\centering
	\includegraphics[width=0.95\textwidth]{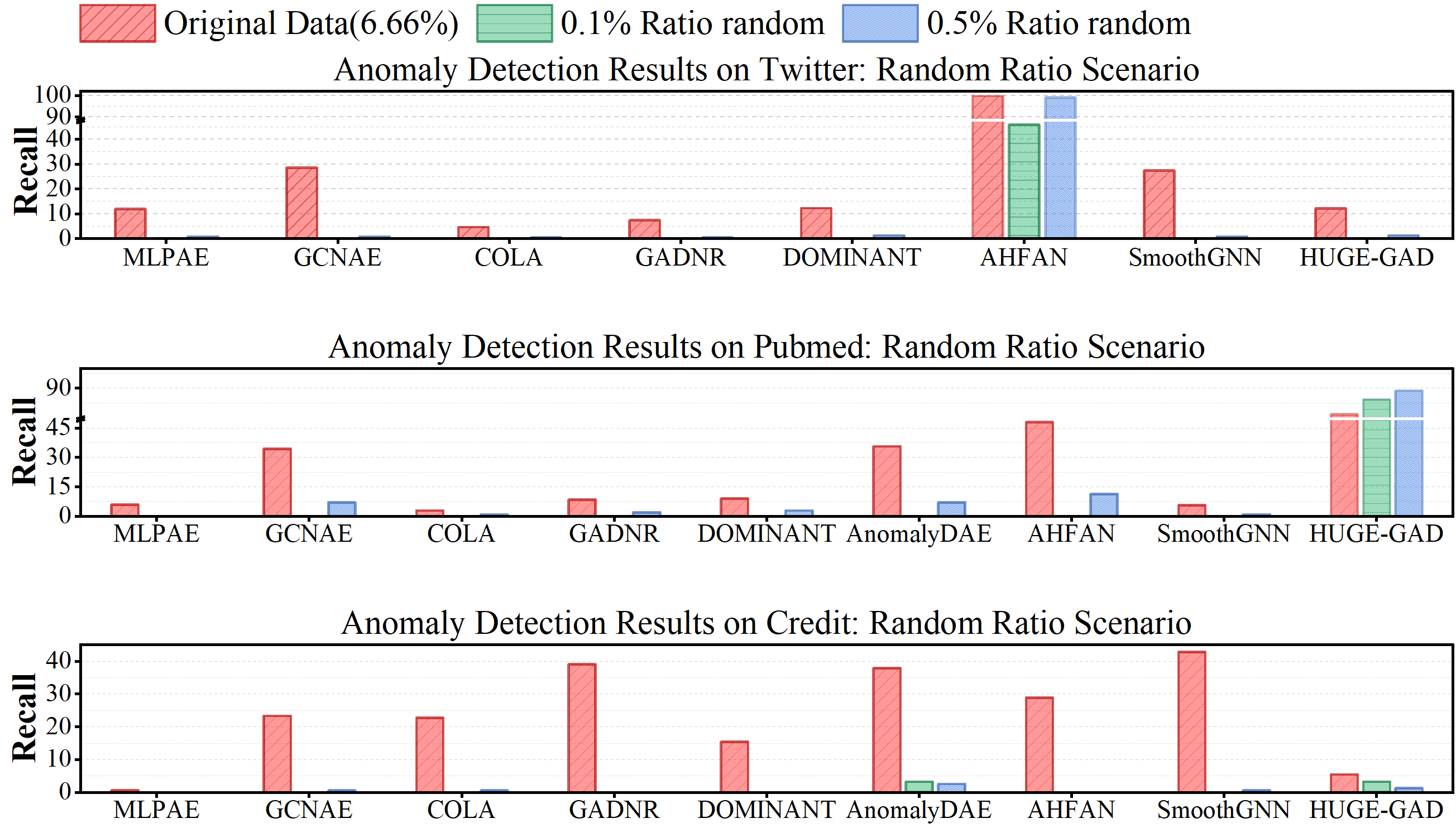}
    \caption{Anomaly detection results across datasets based on varying anomaly ratios.
    Only successfully executed results are shown; models that failed due to OOM are omitted.}
    \label{fig:imbalance}
    \vspace{-1em}
\end{figure*}

 The imbalance evaluation shows that current GAD models are highly sensitive to both anomaly ratio and anomaly distribution pattern.
When the anomaly ratio drops to $0.1\%$, most model--dataset combinations experience a sharp performance decline.
In many low-ratio settings, models such as MLPAE, GCNAE, CoLA, GADNR, DOMINANT, SmoothGNN, and Huge-GAD obtain zero recall, meaning that they fail to detect rare anomalies.
For comprehensive results, please refer to the Appendix~\ref{sec:imbalance}, while specific performance evaluated by Recall is summarized in Figure~\ref{fig:imbalance}.
This result indicates that many existing methods rely implicitly on the assumption that normal patterns dominate the graph while anomalies remain sufficiently frequent to be separable.
Once anomalies become extremely rare, these methods tend to absorb them into the normal distribution and fail to establish an effective anomaly boundary.

\par We also observe a large performance gap across different anomaly distribution patterns.
Core-cluster, edge-cluster, and random-retention settings often lead to substantially different detection results.
For example, on Twitter at the $0.5\%$ anomaly ratio, SmoothGNN achieves $4.62$ Recall under the core-cluster setting, but drops to $0$ under the edge-cluster setting.
AHFAN also shows strong sensitivity to the retention strategy at the $0.1\%$ anomaly ratio, maintaining $100.00$ Recall under core and edge settings but dropping to $45.83$ under random retention.
These results suggest that most current GAD models are sensitive to both anomaly rarity and anomaly distribution pattern.
In other words, strong performance under standard benchmark settings does not necessarily imply robustness under realistic rare-anomaly conditions.

\subsection{Incomplete-Data Analysis: Divergent Effects Across Models and Datasets}
\par
\begin{figure*}[t]
	\centering
	\subfigure {
		\begin{minipage}[b]{0.95\textwidth}
			\centering
			\includegraphics[width=1\textwidth]{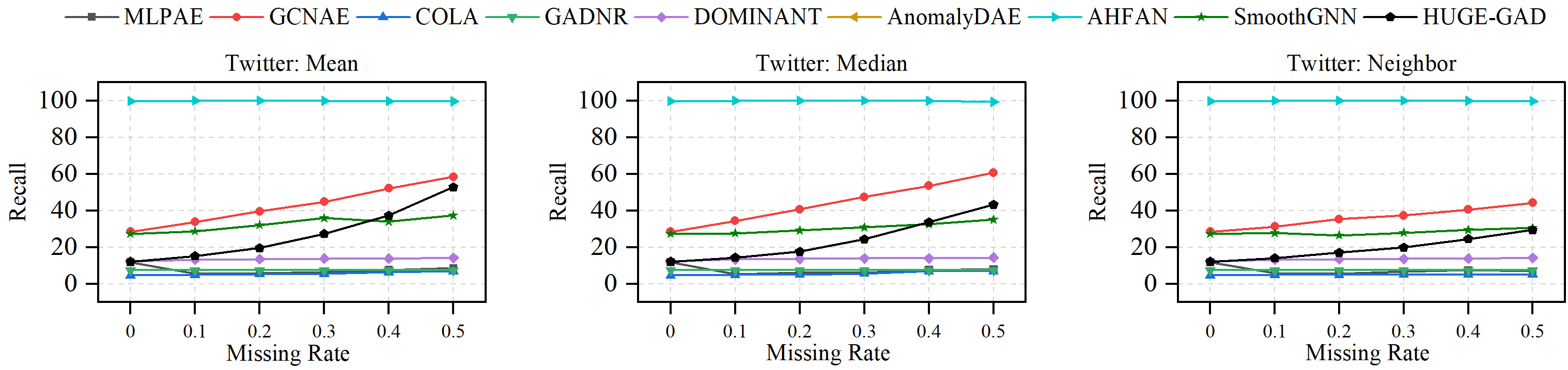}
		\end{minipage}
	}
	\subfigure {
		\begin{minipage}[b]{0.95\textwidth}
			\centering
			\includegraphics[width=1\textwidth]{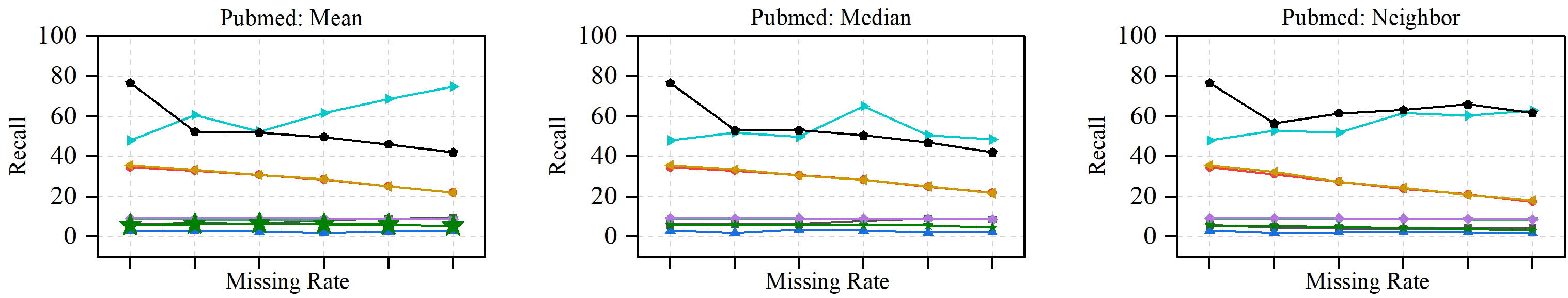}
		\end{minipage}
	}
	\subfigure {
   	\begin{minipage}[b]{0.95\textwidth}
   		\centering
   		\includegraphics[width=1\textwidth]{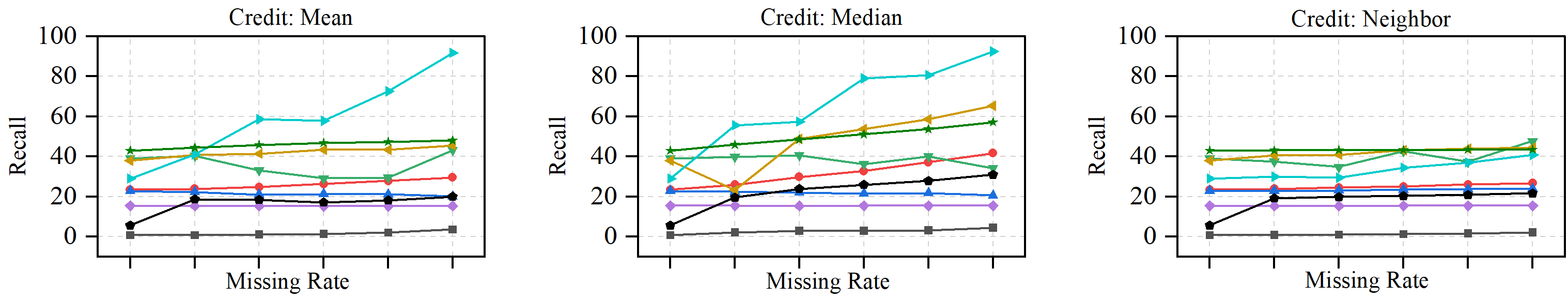}
  	\end{minipage}
  }
  \caption{Anomaly detection results across datasets under varying missing rate.}
  \label{fig:incomplete}
  \vspace{-1em}
\end{figure*}

 The incomplete-data evaluation reveals that attribute missingness does not affect all models in the same way.
Instead, we observe two opposite trends: performance improvement in some settings and performance degradation in others.
For comprehensive results, please refer to the Appendix~\ref{sec:incomplete}, while specific performance evaluated by Recall is summarized in Figure~\ref{fig:incomplete}.

\par On the Credit dataset, several models perform better when attributes are partially missing.
For example, AHFAN improves from $75.93$ AUC-ROC on the complete graph to $99.09$ when the missing ratio reaches $50\%$, corresponding to a gain of $23.16$ points.
By contrast, conventional degradation still appears in many other cases.
Under median imputation, GADNR drops from $52.47$ to $46.34$ as the missing ratio increases to $50\%$.
DOMINANT drops more sharply from $64.30$ to $44.02$ in the same setting.

\par We attribute this divergence to the relative strength of structural and attribute anomaly signals.
When structural anomalies are more informative than attribute anomalies, removing part of the attribute information can reduce distraction and push models toward more effective topological reasoning.
When attribute anomalies are more informative, missing attributes directly remove useful anomaly evidence and cause performance degradation.
For models that already fuse both sources in a tightly coupled way, such as GADNR, missingness mainly acts as information loss and therefore leads to more monotonic degradation.

\par We also find that model sensitivity to imputation choice is dataset- and model-dependent.
For example, on Credit with $50\%$ missing attributes, DOMINANT remains consistently low across mean, median, and neighbor imputation, with AUC-ROC scores of $42.81$, $44.02$, and $43.24$, respectively.
This indicates that incomplete-data robustness should be evaluated jointly over missing ratios and imputation strategies, rather than being summarized by a single filling method.
In addition, reconstruction-based methods are substantially more sensitive to imputation choice than contrastive methods.
For example, DOMINANT exhibits up to a $20.1$-point performance gap across different filling strategies, whereas CoLA varies by less than $4.8$ points.
This indicates that the reported performance of reconstruction-based GAD models can depend heavily on preprocessing choices, while contrastive methods tend to be more stable under incomplete-data conditions.

\section{Limitation}
\par While our benchmark provides a systematic evaluation of GAD models under realistic constraints, several limitations remain.
First, our dataset construction focuses on three fundamental dimensions: scale, class imbalance, and attribute incompleteness. 
We treat these factors as isolated variables to ensure a controlled experimental environment. 
However, real-world deployment scenarios are often significantly more complex and involve a mixture of multiple challenges. 
For instance, a single industrial graph often simultaneously exhibits extreme imbalance, heterophilous structures, and various types of missing data. 
Future versions of this benchmark need to investigate the cross-effects between these factors to better simulate high-noise production environments.

\par Second, our evaluation relies on standard laboratory computational resources. 
While we define a unified hardware protocol to ensure a fair comparison, these resources do not fully represent the massive distributed computing clusters typically found in industry. 
Consequently, our findings regarding the scalability bottleneck reflect model behavior at the laboratory level. 
Evaluation on industrial-grade hardware is necessary to determine whether these limitations can be mitigated through massive hardware acceleration or if they remain inherent to the algorithm designs themselves.

\section{Conclusion and Future Work}
\par In this paper, we presented a multi-dimensional benchmark for GAD to bridge the gap between idealized laboratory settings and the complexities of real-world deployment. 
By systematically evaluating nine representative models across three axes (scalability, class imbalance, and attribute incompleteness), we identified significant limitations in the current state-of-the-art. 
Our empirical study shows that most GNN-based GAD models suffer from severe computational bottlenecks, leading to memory exhaustion or unstable performance when scaled to million-node graphs. 
Furthermore, we observed that existing methods fail to establish effective decision boundaries under realistic low-ratio conditions, often resulting in zero recall for rare instances. 
We also found that while attribute missingness can occasionally drive models to exploit structural signals, reconstruction-based methods remain highly sensitive to the choice of imputation strategy. 
Ultimately, this benchmark provides a diagnostic framework to shift research focus toward building robust and scalable GAD systems for the large-scale, low-quality data encountered in practical applications.

\par Looking forward, several critical directions remain for the development of deployable GAD systems. 
First, there is an urgent need to optimize hardware efficiency, as current methods cannot process industrial-scale datasets under limited computational budgets. 
Enabling algorithms to operate on million-node graphs is a fundamental requirement for practical use. 
Second, while current models struggle at a $0.1\%$ anomaly ratio, real-world data often presents even more extreme cases, such as $0.02\%$. 
Future research should investigate how to effectively identify anomalies when they are exceptionally rare. 
Finally, although general imputation methods allow models to run on incomplete data, GAD requires a unique integration of attribute values and graph topology. 
Therefore, developing specialized, robust GAD methods that natively handle missing information remains a vital task for the community.

\bibliographystyle{plainnat}
\bibliography{reference}

\appendix 

\section{Related Work}

\subsection{Graph Anomaly Detection}
\par GAD~\citep{QiaoTAKAP25,Qing2026iwcmc,KongSWSX25} has received increasing attention due to its importance in applications such as fraud detection, misinformation identification, and risk control~\citep{LiuDZDHZDCPSSLC22,tang2023gadbench}.
Existing methods can be broadly grouped into three lines: reconstruction-based methods, contrastive/self-supervised methods, and methods that incorporate anomaly-specific structural priors.

A major line of work is reconstruction-based GAD, where anomalies are identified through residuals in reconstructing graph structure, node attributes, or both~\citep{KimLBKKYS24,ZhouWLWYYJ24}.
In our benchmark, MLPAE and GCNAE~\citep{KipfW17} serve as simple attribute-only and topology-aware autoencoder baselines.
Representative deep methods in this line include DOMINANT, which jointly reconstructs structure and attributes, and AnomalyDAE, which further models cross-modal interactions through a dual-autoencoder design \citep{ding2019dominant,fan2020anomalydae}.
More recent variants refine the reconstruction target itself.
For example, GAD-NR performs neighborhood reconstruction rather than direct adjacency reconstruction to better capture local anomaly patterns, while AHFAN introduces adaptive attention and graph filtering to improve anomaly modeling under graph heterogeneity \citep{roy2024gadnr,wang2025ahfan}.

Another important line is contrastive and self-supervised GAD~\citep{XuSH0W025}.
Instead of relying only on reconstruction errors, these methods learn anomaly-sensitive representations through contrastive objectives or auxiliary pretext tasks.
CoLA is a representative example that formulates GAD as contrastive self-supervised learning over node--subgraph pairs \citep{liu2022cola}.
FedCAD~\citep{KongZWHCYD25} pseudo‑labels likely anomalies and strengthens contrastive signals by aggregating cross‑client anomaly‑neighbor embeddings, sharpening normal–anomalous separation.
This line is particularly relevant because contrastive methods are often considered more robust than pure reconstruction-based models when graph signals are noisy or distorted.

A third line introduces anomaly-specific structural priors.
These methods argue that anomaly detection benefits from explicitly modeling graph properties that standard message passing may overlook.
For example, ComGA incorporates community-aware signals for characterizing structural anomalies \citep{luo2022comga}, while SmoothGNN exploits the distinct smoothing behavior of anomalous nodes \citep{dong2025smoothgnn}.
In fraud-oriented scenarios, HUGE further leverages heterophily-guided anomaly modeling \citep{pan2025huge}.
Together, these works reflect a broader trend from generic graph encoders toward anomaly-oriented inductive biases.

Recent work has also begun to revisit GAD evaluation from a benchmark perspective.
In particular, GADBench provides a systematic benchmark for supervised and semi-supervised graph anomaly detection, highlighting the importance of standardized evaluation and large-scale efficiency analysis \citep{tang2023gadbench}.
However, existing benchmark efforts still focus primarily on supervised settings and do not explicitly study several realism gaps that are central to practical unsupervised GAD, including anomaly rarity, incomplete-data robustness, and controlled data variants under a strict variable-isolation protocol.
As a result, prior work has focused much more on improving detector architectures than on systematically characterizing when GAD models remain reliable under deployment-oriented conditions.
This realism gap motivates the present work.

\subsection{Dataset Construction and Variant Beyond GAD}

Although controlled dataset variant has not been systematically developed for GAD, related ideas have been explored in the broader graph learning literature~\citep{XiaPRFLSYK25}.
These works are relevant because they show how graph benchmarks can move beyond evaluation on a few static datasets and instead expose model limitations through controlled benchmark construction and stress testing.

One relevant direction is controllable benchmark construction.
Frameworks such as GraphWorld, GOOD, and GRB demonstrate the value of controllable graph generation, structured distribution shifts, and unified robustness evaluation for diagnosing model behavior under non-standard testing conditions \citep{palowitch2022graphworld,gui2022good,zheng2021grb}.
These studies provide an important benchmark-design perspective, but they target general graph learning problems such as out-of-distribution generalization or adversarial robustness, rather than node-level anomaly detection.

Another related direction is imbalance-aware graph evaluation.
\textbf{IGL-Bench} shows that imbalance itself can serve as a first-class benchmark dimension in graph learning and enables unified analysis of model behavior under imbalanced settings \citep{qin2025iglbench}.
This insight is conceptually related to our imbalance robustness evaluation, since anomaly detection is inherently tied to rarity.
However, existing imbalance benchmarks are mainly designed for supervised class imbalance, whereas GAD focuses on rare anomalous instances whose prevalence and distribution patterns differ fundamentally from standard minority classes.

A third relevant direction studies graphs with incomplete node attributes.
Representative works include GRAPE, which formulates missing-data handling through graph representation learning \citep{you2020grape}, Feature Propagation, which shows that simple propagation-based imputation can be highly effective and scalable \citep{rossi2022featureprop}, and confidence-aware imputation methods such as PCFI \citep{um2023pcfi}.
These works provide practical mechanisms for constructing graph variants with controlled missingness and feature repair.
However, they are mainly developed for tasks such as node classification or feature recovery, rather than anomaly detection, and therefore do not examine incomplete-data robustness in GAD, namely how missing attributes alter anomaly signals or how imputation strategies affect anomaly ranking.

In summary, prior work outside GAD has established several useful principles for data-centric graph evaluation, including controllable benchmark construction, imbalance-aware testing, and controlled missing-data simulation with imputation.
Our work builds on these general ideas, but adapts them to the specific realism gap of GAD by jointly studying scalability, imbalance robustness, and incomplete-data robustness under a unified variable-isolation protocol.

\section{Dataset Details and Prepocessing} \label{sec:data}
\subsection{Dataset Descriptions}
We select diverse datasets to ensure our benchmark covers various graph scales, attribute dimensions, and anomaly types:
\begin{itemize} [leftmargin=0.5cm]
    \item \textbf{Twitter}~\citep{neo2024fairgad}: A social interaction graph used to identify COVID-$19$ misinformation spreaders. It comprises $47,712$ nodes and $468,697$ edges. 
    Nodes represent users with 780-dimensional features derived from profiles and posting behaviors; the anomaly ratio is $6.66$\%.
    \item \textbf{PubMed}~\citep{sen2008collective}: A biomedical citation network where nodes are papers and edges represent citations. 
    Anomalies ($3.04$\%) are papers with structural or thematic deviations. 
    It contains $19,717$ nodes and $500$-dimensional TF-IDF features.
    \item \textbf{Credit}~\citep{yeh2009credit}: A user-behavior graph for credit risk analysis. Nodes represent $30,000$ credit card users, and edges reflect high behavioral similarity. 
    Anomalies ($22.12$\%) represent default users based on $23$-dimensional repayment and consumption records.
    \item \textbf{DGraph-Fin}~\citep{huang2022dgraph}: An industrial-grade financial transaction graph for fraud detection. 
    It contains over $3.7$ million nodes and $4.3$ million edges. 
    Anomalies ($1.27$\%) are confirmed fraudulent accounts identified through $17$-dimensional identity and transaction features.
    \item \textbf{T-Social}~\citep{tang2022rethinking}: A super-large-scale social graph containing $5.7$ million nodes and $73$ million edges. 
    It targets users violating platform rules ($3.01$\% anomaly ratio) using $10$-dimensional social behavior attributes.
\end{itemize}

\subsection{Standardized Preprocessing Pipeline}
\par To ensure a fair comparison across all GAD models, we implement a rigorous three-step preprocessing workflow:
Firstly, we convert all raw data into a standardized format, eliminating isolated nodes, duplicate edges, and self-loops to ensure the graph structure is clean and connected.
Secondly, all node features are converted to $32$-bit floating-point tensors. 
We apply min-max normalization to each dimension to prevent features with large scales from dominating the model training process.
Thirdly, we perform a consistency check to ensure every node has a unique, binary anomaly label. 
This step removes any potential noise from missing or duplicate annotations in the original source files.

\section{Hardware and Software Environment} \label{sec:hardware}
\par All experiments were conducted on a high-performance computing cluster to ensure efficiency metrics (training time and memory) are directly comparable.
CPU: Intel(R) Xeon(R) Gold $6455$B.
GPU: NVIDIA L$40$-$48$GB.
Software: Ubuntu $22.04$, Python $3.8.20$, PyTorch $2.1.0$+cu$121$, and PyG (PyTorch Geometric) $2.6.1$.
The hardware specifications of all evaluated baselines are summarized and presented in Table~\ref{tab:hardware}.

\begin{table}[t]
    \centering
    \renewcommand{\arraystretch}{1.25}
    \caption{Hardware specifications of the evaluated baselines. "N/A" denotes that the corresponding information is not explicitly provided in the original publication.}
    \footnotesize
    \begin{tabular}{lrr}
      \toprule
      \textbf{Methods} & \textbf{Hardware} & \textbf{Memory} \\
      \midrule
      MLPAE & N/A   & N/A \\
      GCNAE~\citep{KipfW17} & NVIDIA GTX Titan X      & $12$ GB \\
      COLA~\citep{liu2022cola} & NVIDIA RTX $2070$      & $8$ GB \\
      GADNR~\citep{roy2024gadnr} & NVIDIA A$10$   & $24$ GB \\
      DOMINANT~\citep{ding2019dominant} & N/A   & N/A \\
      AnomalyDAE~\citep{fan2020anomalydae} & N/A   & N/A \\
      AHFAN~\citep{wang2025ahfan} & Tesla A$40$     & $48$ GB \\
      SmoothGNN~\citep{dong2025smoothgnn} & N/A   & N/A \\
      HUGE-GAD~\cite{pan2025huge} & N/A   & $24$ GB \\
      Our Benchmark & NVIDIA L$40$ & $48$ GB\\
      \bottomrule
    \end{tabular}
  \label{tab:hardware}
\end{table}

\section{Baseline Model Descriptions} \label{sec:baseline}
We provide a brief overview of the evaluated GAD methods:
\begin{itemize} [leftmargin=0.5cm]
    \item MLPAE: A Multi-Layer Perceptron Autoencoder that learns to reconstruct node attributes. 
    It serves as a structural-agnostic baseline.
    \item GCNAE~\citep{KipfW17}: A Graph Convolutional Network Autoencoder that uses GCN layers to encode both structure and attributes into a latent space.
    \item CoLA~\citep{liu2022cola}: Uses contrastive learning by comparing a node's local subgraph with a sampled negative instance.
    \item GADNR~\citep{roy2024gadnr}: A reconstruction-based model that predicts the neighborhood degree distribution and attribute concentration.
    \item DOMINANT~\citep{ding2019dominant}: A benchmark dual-autoencoder that uses a GCN encoder and separate decoders for attribute and structure reconstruction.
    \item AnomalyDAE~\citep{fan2020anomalydae}: Captures cross-modal correlations using an attention-based encoder to weight the importance of neighbors.
    \item AHFAN~\citep{wang2025ahfan}: An adaptive hypergraph-based model that uses frequency-domain filtering to identify anomalies.
    \item SmoothGNN~\citep{dong2025smoothgnn}: A recent method that utilizes the "smoothing difficulty" of nodes to distinguish anomalies from normal nodes efficiently.
    \item Huge-GAD~\cite{pan2025huge}: Specifically designed for large-scale graphs, it leverages label-free heterophily signatures to guide detection.
\end{itemize}

\section{Results}
\subsection{Results of Scalability Analysis} \label{sec:scalability}
\begin{table}[H] 
  \centering
  \footnotesize
  \setlength{\tabcolsep}{12pt}
  \caption{Anomaly Detection Results on DGraph-Fin and T-Social Dataset}
  \begin{tabular}{lrrrrr}
    \toprule
    \textbf{Algorithm} & \textbf{AUCROC} & \textbf{AUCPRC} & \textbf{Recall} & \textbf{Runtime (s)} & \textbf{Memory (MB)} \\
    \midrule
    \rowcolor{red!20} \multicolumn{6}{c}{\textbf{DGraph-Fin ($3700$k Nodes)}} \\
    MLPAE & $45.29$ & $0.36$ & $0.04$ & $369.97$ & $9,074.00$ \\
    GCNAE & $53.80$ & $0.34$ & $0.14$ & $494.67$ & $11,666.00$ \\
    COLA  & $50.37$ & $0.42$ & $0.07$ & $560.83$ & $16,478$ \\
    GADNR & OOM & OOM & OOM & OOM & OOM \\
    DOMINANT & OOM & OOM & OOM & OOM & OOM \\
    AnomalyDAE & OOM & OOM & OOM & OOM & OOM \\
    AHFAN & $64.00$ & $0.65$ & $0.00$ & $1,913.98$ & $21,392.00$  \\
    SmoothGNN & $51.37$ & $0.43$ & $0.40$ & $318.21$ & $25,693.43$  \\
    HUGE-GAD & OOM & OOM & OOM & OOM & OOM  \\
    \midrule
    \rowcolor{red!20} \multicolumn{6}{c}{\textbf{T-Social ($5781$k Nodes)}} \\
    MLPAE & $39.98$ & $2.34$ & $1.56$ & $633.42$ & $18,312.00$ \\
    GCNAE & OOM & OOM & OOM & OOM & OOM \\
    COLA & OOM & OOM & OOM & OOM & OOM \\
    GADNR & OOM & OOM & OOM & OOM & OOM \\
    DOMINANT & OOM & OOM & OOM & OOM & OOM \\
    AnomalyDAE & OOM & OOM & OOM & OOM & OOM \\
    AHFAN & OOM & OOM & OOM & OOM & OOM  \\
    SmoothGNN & OOM & OOM & OOM & OOM & OOM  \\
    HUGE-GAD & OOM & OOM & OOM & OOM & OOM  \\ 
    \bottomrule
  \end{tabular}
  \label{tab:dgraphfin_expanded}
\end{table}

\begin{table}[H]
  \centering
  \footnotesize
  \setlength{\tabcolsep}{12pt}
  \caption{Anomaly Detection Results on Twitter and Pubmed Expanded Dataset (Grouped by Scale)}
    \begin{tabular}{lrrrrr}
      \toprule
      \textbf{Algorithm} & \textbf{AUCROC} & \textbf{AUCPRC} & \textbf{Recall} & \textbf{Runtime (s)} & \textbf{Memory (MB)} \\
      \midrule
      \rowcolor{red!20} \multicolumn{6}{c}{\textbf{Original Twitter Graph ($47$k Nodes)}} \\
      MLPAE      & $85.42$ & $19.59$ & $11.87$ & $5.00$   & $970.00$ \\
      GCNAE      & $91.19$ & $28.33$ & $28.52$ & $13.22$  & $4,106.00$ \\
      COLA       & $61.76$ & $6.17$  & $4.63$  & $9.08$   & $1,032.00$ \\
      GADNR      & $57.42$ & $7.92$  & $7.46$  & $212.58$ & $18,616.00$ \\
      DOMINANT   & $69.71$ & $10.45$ & $12.24$ & $147.44$ & $44,510.00$ \\
      AnomalyDAE & OOM     & OOM     & OOM     & OOM      & OOM \\
      AHFAN      & $100.00$& $100.00$& $99.68$ & $32.81$  & $1,320.00$ \\
      SmoothGNN  & $79.85$ & $21.08$ & $27.38$ & $9.01$   & $4,515.93$ \\
      HUGE-GAD   & $63.01$ & $9.91$  & $12.18$ & $78.86$  & $12,989.00$ \\
      \cmidrule(lr){1-6}
      \rowcolor{red!20} \multicolumn{6}{c}{\textbf{Scale-Expanded Twitter Graph ($500$k Nodes)}} \\
      MLPAE      & $84.94$ & $19.14$ & $12.01$ & $52.60$  & $8,626.00$ \\
      GCNAE      & $88.31$ & $25.43$ & $28.58$ & $190.10$ & $42,852.00$ \\
      COLA       & $54.90$ & $6.20$  & $5.76$  & $126.61$ & $10,644.00$ \\
      GADNR      & OOM     & OOM     & OOM     & OOM      & OOM \\
      DOMINANT   & OOM     & OOM     & OOM     & OOM      & OOM \\
      AnomalyDAE & OOM     & OOM     & OOM     & OOM      & OOM \\
      AHFAN      & $100.00$& $100.00$& $100.00$& $407.51$ & $10,663.00$ \\
      SmoothGNN  & $76.56$ & $12.80$ & $6.88$  & $54.41$  & $30,530.95$ \\
      HUGE-GAD   & OOM     & OOM     & OOM     & OOM      & OOM \\
      \cmidrule(lr){1-6}
      \rowcolor{red!20} \multicolumn{6}{c}{\textbf{Scale-Expanded Twitter Graph ($1,000$k Nodes)}} \\
      MLPAE      & $84.85$ & $19.21$ & $12.42$ & $110.33$ & $17,212.00$ \\
      GCNAE      & OOM     & OOM     & OOM     & OOM      & OOM \\
      COLA       & $54.81$ & $6.14$  & $5.33$  & $270.90$ & $21,254.00$ \\
      GADNR      & OOM     & OOM     & OOM     & OOM      & OOM \\
      DOMINANT   & OOM     & OOM     & OOM     & OOM      & OOM \\
      AnomalyDAE & OOM     & OOM     & OOM     & OOM      & OOM \\
      AHFAN      & $100.00$& $99.98$ & $100.00$& $654.04$ & $21,300.00$ \\
      SmoothGNN  & OOM     & OOM     & OOM     & OOM      & OOM \\
      HUGE-GAD   & OOM     & OOM     & OOM     & OOM      & OOM \\
      \midrule
      
      \rowcolor{red!20} \multicolumn{6}{c}{\textbf{Original Pubmed Graph ($19$k Nodes)}} \\
      MLPAE      & $73.07$ & $11.04$ & $6.00$  & $6.05$   & $330.00$ \\
      GCNAE      & $74.41$ & $23.37$ & $34.50$ & $4.69$   & $714.00$ \\
      COLA       & $49.81$ & $3.18$  & $3.00$  & $3.49$   & $320.00$ \\
      GADNR      & $75.75$ & $9.21$  & $8.50$  & $103.67$ & $7,532.00$ \\
      DOMINANT   & $81.26$ & $10.16$ & $9.17$  & $16.73$  & $10,502.00$ \\
      AnomalyDAE & $74.82$ & $23.90$ & $35.67$ & $17.92$  & $12,072.00$ \\
      AHFAN      & $81.24$ & $25.38$ & $48.03$ & $14.65$  & $265.00$ \\
      SmoothGNN  & $52.63$ & $3.76$  & $5.67$  & $3.69$   & $1,929.07$ \\
      HUGE-GAD   & $94.73$ & $55.11$ & $76.67$ & $63.36$  & $4,946.00$ \\
      \cmidrule(lr){1-6} 

      \rowcolor{red!20} \multicolumn{6}{c}{\textbf{Scale-Expanded Pubmed Graph ($500$k Nodes)}} \\
      MLPAE      & $79.98$ & $11.27$ & $0.30$  & $53.90$  & $5,942.00$ \\
      GCNAE      & $98.61$ & $77.94$ & $76.75$ & $108.09$ & $17,690.00$ \\
      COLA       & $50.48$ & $3.54$  & $0.99$  & $89.81$  & $7,114.00$ \\
      GADNR      & OOM     & OOM     & OOM     & OOM      & OOM \\
      DOMINANT   & OOM     & OOM     & OOM     & OOM      & OOM \\
      AnomalyDAE & OOM     & OOM     & OOM     & OOM      & OOM \\
      AHFAN      & $97.94$ & $95.90$ & $89.66$ & $237.33$ & $6,154.00$ \\
      SmoothGNN  & $84.40$ & $13.59$ & $3.49$  & $48.24$  & $20,219.83$ \\
      HUGE-GAD   & OOM     & OOM     & OOM     & OOM      & OOM \\
      \cmidrule(lr){1-6} 

      \rowcolor{red!20} \multicolumn{6}{c}{\textbf{Scale-Expanded Pubmed Graph ($1,000$k Nodes)}} \\
      MLPAE      & $80.26$ & $10.94$ & $0.28$  & $106.19$ & $11,842.00$ \\
      GCNAE      & $98.90$ & $83.11$ & $79.00$ & $240.96$ & $35,418.00$ \\
      COLA       & $51.02$ & $3.66$  & $0.64$  & $202.45$ & $14,194.00$ \\
      GADNR      & OOM     & OOM     & OOM     & OOM      & OOM \\
      DOMINANT   & OOM     & OOM     & OOM     & OOM      & OOM \\
      AnomalyDAE & OOM     & OOM     & OOM     & OOM      & OOM \\
      AHFAN      & $99.55$ & $98.48$ & $91.38$ & $474.87$ & $12,293.00$ \\
      SmoothGNN  & $86.48$ & $19.30$ & $30.21$ & $99.56$  & $40,375.34$ \\
      HUGE-GAD   & OOM     & OOM     & OOM     & OOM      & OOM \\
      \bottomrule
    \end{tabular}
  \label{tab:twitter_expanded}
\end{table}

\begin{table}[H]
  \centering
  \footnotesize
  \setlength{\tabcolsep}{12pt}
  \caption{Anomaly Detection Results on Credit Expanded Dataset (Grouped by Scale)}
    \begin{tabular}{lrrrrr}
      \toprule
      \textbf{Algorithm} & \textbf{AUCROC} & \textbf{AUCPRC} & \textbf{Recall} & \textbf{Runtime (s)} & \textbf{Memory (MB)} \\
      \cmidrule(lr){1-6} 
      
      \rowcolor{red!20} \multicolumn{6}{c}{\textbf{Original Credit Graph ($30$k Nodes)}} \\
      MLPAE      & $58.73$ & $25.30$ & $0.71$  & $3.08$   & $226.00$ \\
      GCNAE      & $66.74$ & $25.26$ & $23.40$ & $23.63$  & $2,416.00$ \\
      COLA       & $52.91$ & $23.48$ & $22.78$ & $25.45$  & $2,374.00$ \\
      GADNR      & $52.47$ & $23.63$ & $39.12$ & $462.04$ & $14,498.00$ \\
      DOMINANT   & $64.30$ & $18.76$ & $15.45$ & $50.93$  & $23,084.00$ \\
      AnomalyDAE & $66.83$ & $34.00$ & $37.94$ & $51.90$  & $25,776.00$ \\
      AHFAN      & $75.93$ & $46.74$ & $28.93$ & $90.49$  & $4,528.00$ \\
      SmoothGNN  & $70.38$ & $38.08$ & $42.90$ & $4.21$   & $2,589.84$ \\
      HUGE-GAD   & $52.68$ & $22.22$ & $5.56$  & $203.17$ & $12,792.00$ \\
      \cmidrule(lr){1-6} 

      \rowcolor{red!20} \multicolumn{6}{c}{\textbf{Scale-Expanded Credit Graph ($500$k Nodes)}} \\
      MLPAE      & $58.11$ & $25.33$ & $3.12$  & $48.52$  & $3,018.00$ \\
      GCNAE      & $61.52$ & $28.16$ & $23.62$ & $663.20$ & $39,562.00$ \\
      COLA       & $51.24$ & $22.43$ & $13.96$ & $694.40$ & $39,554.00$ \\
      GADNR      & OOM     & OOM     & OOM     & OOM      & OOM \\
      DOMINANT   & OOM     & OOM     & OOM     & OOM      & OOM \\
      AnomalyDAE & OOM     & OOM     & OOM     & OOM      & OOM \\
      AHFAN      & OOM     & OOM     & OOM     & OOM      & OOM \\
      SmoothGNN  & $69.73$ & $37.48$ & $40.70$ & $42.38$  & $6,494.14$ \\
      HUGE-GAD   & OOM     & OOM     & OOM     & OOM      & OOM \\
      \cmidrule(lr){1-6} 

      \rowcolor{red!20} \multicolumn{6}{c}{\textbf{Scale-Expanded Credit Graph ($1,000$k Nodes)}} \\
      MLPAE      & $58.20$ & $25.31$ & $3.11$  & $101.28$ & $6,012.00$ \\
      GCNAE      & OOM     & OOM     & OOM     & OOM      & OOM \\
      COLA       & OOM     & OOM     & OOM     & OOM      & OOM \\
      GADNR      & OOM     & OOM     & OOM     & OOM      & OOM \\
      DOMINANT   & OOM     & OOM     & OOM     & OOM      & OOM \\
      AnomalyDAE & OOM     & OOM     & OOM     & OOM      & OOM \\
      AHFAN      & OOM     & OOM     & OOM     & OOM      & OOM \\
      SmoothGNN  & $70.38$ & $37.66$ & $41.41$ & $84.53$  & $12,916.14$ \\
      HUGE-GAD   & OOM     & OOM     & OOM     & OOM      & OOM \\
      \bottomrule
    \end{tabular}
  \label{tab:credit_expanded}
\end{table}
\subsection{Results of Imbalance Analysis} \label{sec:imbalance}
\begin{table}[H]
    \centering
    \footnotesize
    \caption{Performance Comparison across Different Anomaly Ratios on Twitter. ROC, PRC, and Rec denote AUC-ROC, AUC-PRC, and Recall, respectively.}
    \begin{tabular}{lrrrrrrrrr}
        \toprule
        \textbf{Algorithm} & \textbf{ROC} & \textbf{PRC} & \textbf{Rec} & \textbf{ROC} & \textbf{PRC} & \textbf{Rec} & \textbf{ROC} & \textbf{PRC} & \textbf{Rec} \\
        \midrule
        & \multicolumn{3}{c}{\cellcolor{green!20} \textbf{Twitter ($6.66\%$ Anomaly Ratio)}} & \multicolumn{3}{c}{\cellcolor{blue!20} \textbf{$0.1\%$ Ratio Core}} & \multicolumn{3}{c}{\cellcolor{yellow!20} \textbf{$0.5\%$ Ratio Core}}\\
        MLPAE      & $85.42$ & $19.59$ & $11.87$ & $87.32$ & $0.28$  & $0.00$  & $86.63$ & $1.40$  & $0.00$  \\
        GCNAE      & $91.19$ & $28.33$ & $28.52$ & $93.67$ & $0.66$  & $0.00$  & $93.54$ & $3.13$  & $0.00$  \\
        COLA       & $61.76$ & $6.17$  & $4.63$  & $66.50$ & $0.10$  & $0.00$  & $66.44$ & $0.45$  & $0.42$  \\
        GADNR      & $57.42$ & $7.92$  & $7.46$  & $61.99$ & $0.14$  & $0.00$  & $63.49$ & $0.70$  & $0.84$  \\
        DOMINANT   & $69.71$ & $10.45$ & $12.24$ & $71.91$ & $0.26$  & $2.13$  & $72.71$ & $0.99$  & $1.26$  \\
        AnomalyDAE & OOM     & OOM     & OOM     & OOM     & OOM     & OOM     & OOM     & OOM     & OOM     \\
        AHFAN      & $100.00$& $100.00$& $99.68$ & $100.00$& $100.00$& $100.00$& $100.00$& $100.00$& $100.00$\\
        SmoothGNN  & $79.85$ & $21.08$ & $27.38$ & $77.53$ & $0.30$  & $0.00$  & $75.95$ & $1.67$  & $4.62$  \\
        HUGE-GAD   & $63.01$ & $9.91$  & $12.18$ & $44.32$ & $0.07$  & $0.00$  & $42.82$ & $0.39$  & $0.00$  \\
        \midrule

        & \multicolumn{3}{c}{\cellcolor{green!20} \textbf{Twitter ($6.66\%$ Anomaly Ratio)}} & \multicolumn{3}{c}{\cellcolor{blue!20} \textbf{$0.1\%$ Ratio edge}} & \multicolumn{3}{c}{\cellcolor{yellow!20} \textbf{$0.5\%$ Ratio edge}}\\
        MLPAE      & $85.42$ & $19.59$ & $11.87$ & $90.19$ & $0.45$  & $0.00$  & $91.50$ & $2.62$  & $0.00$  \\
        GCNAE      & $91.19$ & $28.33$ & $28.52$ & $91.33$ & $0.53$  & $0.00$  & $92.92$ & $3.18$  & $0.00$  \\
        COLA       & $61.76$ & $6.17$  & $4.63$  & $68.64$ & $0.12$  & $0.00$  & $65.21$ & $0.52$  & $0.42$  \\
        GADNR      & $57.42$ & $7.92$  & $7.46$  & $70.19$ & $0.17$  & $0.00$  & $66.96$ & $0.80$  & $0.42$  \\
        DOMINANT   & $69.71$ & $10.45$ & $12.24$ & $82.75$ & $0.28$  & $0.00$  & $78.40$ & $1.34$  & $1.26$  \\
        AnomalyDAE & OOM     & OOM     & OOM     & OOM     & OOM     & OOM     & OOM     & OOM     & OOM     \\
        AHFAN      & $100.00$& $100.00$& $99.68$ & $100.00$& $100.00$& $100.00$& $100.00$& $98.52$ & $100.00$\\
        SmoothGNN  & $79.85$ & $21.08$ & $27.38$ & $61.48$ & $0.12$  & $0.00$  & $61.71$ & $0.60$  & $0.00$  \\
        HUGE-GAD   & $63.01$ & $9.91$  & $12.18$ & $79.94$ & $0.23$  & $0.00$  & $78.77$ & $1.28$  & $0.00$  \\
        \bottomrule
    \end{tabular}
    \label{tab:twitter_pubmed_comparison}
\end{table}

\begin{table}[H]
    \centering
    \footnotesize
    \caption{Performance Comparison across Different Anomaly Ratios on Twitter, Pubmed and Credit. ROC, PRC, and Rec denote AUC-ROC, AUC-PRC, and Recall, respectively.}
    \begin{tabular}{lrrrrrrrrr}
        \toprule
        \textbf{Algorithm} & \textbf{ROC} & \textbf{PRC} & \textbf{Rec} & \textbf{ROC} & \textbf{PRC} & \textbf{Rec} & \textbf{ROC} & \textbf{PRC} & \textbf{Rec} \\
        \midrule
        & \multicolumn{3}{c}{\cellcolor{green!20} \textbf{Twitter ($6.66\%$ Anomaly Ratio)}} & \multicolumn{3}{c}{\cellcolor{blue!20} \textbf{$0.1\%$ Ratio random}} & \multicolumn{3}{c}{\cellcolor{yellow!20} \textbf{$0.5\%$ Ratio random}}\\
        MLPAE      & $85.42$ & $19.59$ & $11.87$ & $89.14$ & $0.44$  & $0.00$  & $88.61$ & $2.11$  & $0.84$  \\
        GCNAE      & $91.19$ & $28.33$ & $28.52$ & $91.66$ & $0.68$  & $0.00$  & $92.34$ & $3.26$  & $0.84$  \\
        COLA       & $61.76$ & $6.17$  & $4.63$  & $66.00$ & $0.10$  & $0.00$  & $61.90$ & $0.49$  & $0.42$  \\
        GADNR      & $57.42$ & $7.92$  & $7.46$  & $61.10$ & $0.14$  & $0.00$  & $55.27$ & $0.57$  & $0.42$  \\
        DOMINANT   & $69.71$ & $10.45$ & $12.24$ & $74.95$ & $0.20$  & $0.00$  & $70.58$ & $0.86$  & $1.26$  \\
        AnomalyDAE & OOM     & OOM     & OOM     & OOM     & OOM     & OOM     & OOM     & OOM     & OOM     \\
        AHFAN      & $100.00$& $100.00$& $99.68$ & $94.16$ & $80.58$ & $45.83$ & $100.00$& $100.00$& $99.04$ \\
        SmoothGNN  & $79.85$ & $21.08$ & $27.38$ & $71.16$ & $0.25$  & $0.00$  & $70.33$ & $1.08$  & $0.84$  \\
        HUGE-GAD   & $63.01$ & $9.91$  & $12.18$ & $61.45$ & $0.13$  & $0.00$  & $57.88$ & $0.63$  & $1.26$  \\
        \midrule

        & \multicolumn{3}{c}{\cellcolor{green!20} \textbf{Pubmed ($3.04\%$ Anomaly Ratio)}} & \multicolumn{3}{c}{\cellcolor{blue!20} \textbf{$0.1\%$ Ratio Core}} & \multicolumn{3}{c}{\cellcolor{yellow!20} \textbf{$0.5\%$ Ratio Core}}\\
        MLPAE      & $73.07$ & $11.04$ & $6.00$  & $85.31$ & $0.06$  & $0.00$  & $75.29$ & $0.32$  & $0.00$  \\
        GCNAE      & $74.41$ & $23.37$ & $34.50$ & $3.35$  & $0.05$  & $0.00$  & $16.39$ & $0.28$  & $0.00$  \\
        COLA       & $49.81$ & $3.18$  & $3.00$  & $80.33$ & $0.23$  & $0.00$  & $70.75$ & $0.91$  & $0.00$  \\
        GADNR      & $75.75$ & $9.21$  & $8.50$  & $93.27$ & $0.70$  & $0.00$  & $93.84$ & $3.58$  & $0.00$  \\
        DOMINANT   & $81.26$ & $10.16$ & $9.17$  & $91.99$ & $0.58$  & $0.00$  & $92.89$ & $3.09$  & $0.00$  \\
        AnomalyDAE & $74.82$ & $23.90$ & $35.67$ & $15.12$ & $0.05$  & $0.00$  & $24.73$ & $0.28$  & $0.00$  \\
        AHFAN      & $81.24$ & $25.38$ & $48.03$ & $48.65$ & $0.08$  & $0.00$  & $76.84$ & $4.42$  & $8.33$  \\
        SmoothGNN  & $52.63$ & $3.76$  & $5.67$  & $53.69$ & $0.10$  & $0.00$  & $52.21$ & $0.50$  & $1.02$  \\
        HUGE-GAD   & $94.73$ & $55.11$ & $76.67$ & $97.27$ & $3.31$  & $94.74$ & $97.31$ & $14.45$ & $87.76$ \\
        \midrule

        & \multicolumn{3}{c}{\cellcolor{green!20} \textbf{Pubmed ($3.04\%$ Anomaly Ratio)}} & \multicolumn{3}{c}{\cellcolor{blue!20} \textbf{$0.1\%$ Ratio edge}} & \multicolumn{3}{c}{\cellcolor{yellow!20} \textbf{$0.5\%$ Ratio edge}}\\
        MLPAE      & $73.07$ & $11.04$ & $6.00$  & $97.58$ & $2.07$  & $0.00$  & $97.81$ & $12.19$ & $4.08$  \\
        GCNAE      & $74.41$ & $23.37$ & $34.50$ & $98.66$ & $7.93$  & $0.00$  & $99.40$ & $44.64$ & $45.92$ \\
        COLA       & $49.81$ & $3.18$  & $3.00$  & $84.14$ & $0.32$  & $0.00$  & $58.80$ & $0.47$  & $0.00$  \\
        GADNR      & $75.75$ & $9.21$  & $8.50$  & $61.33$ & $0.13$  & $0.00$  & $61.39$ & $0.68$  & $0.00$  \\
        DOMINANT   & $81.26$ & $10.16$ & $9.17$  & $71.23$ & $0.18$  & $0.00$  & $74.09$ & $1.02$  & $1.02$  \\
        AnomalyDAE & $74.82$ & $23.90$ & $35.67$ & $98.77$ & $7.66$  & $0.00$  & $99.43$ & $45.90$ & $46.94$ \\
        AHFAN      & $81.24$ & $25.38$ & $48.03$ & $88.17$ & $40.63$ & $33.33$ & $74.36$ & $20.28$ & $38.46$ \\
        SmoothGNN  & $52.63$ & $3.76$  & $5.67$  & $66.74$ & $0.84$  & $0.00$  & $49.92$ & $1.20$  & $4.08$  \\
        HUGE-GAD   & $94.73$ & $55.11$ & $76.67$ & $94.07$ & $3.05$  & $73.68$ & $91.75$ & $27.77$ & $73.47$ \\
        \midrule

        & \multicolumn{3}{c}{\cellcolor{green!20} \textbf{Pubmed ($3.04\%$ Anomaly Ratio)}} & \multicolumn{3}{c}{\cellcolor{blue!20} \textbf{$0.1\%$ Ratio random}} & \multicolumn{3}{c}{\cellcolor{yellow!20} \textbf{$0.5\%$ Ratio random}}\\
        MLPAE      & $73.07$ & $11.04$ & $6.00$  & $76.47$ & $0.34$  & $0.00$  & $71.82$ & $1.68$  & $0.00$  \\
        GCNAE      & $74.41$ & $23.37$ & $34.50$ & $71.49$ & $0.81$  & $0.00$  & $73.34$ & $3.97$  & $7.14$  \\
        COLA       & $49.81$ & $3.18$  & $3.00$  & $67.42$ & $0.27$  & $0.00$  & $50.62$ & $0.57$  & $1.02$  \\
        GADNR      & $75.75$ & $9.21$  & $8.50$  & $86.69$ & $0.61$  & $0.00$  & $79.41$ & $1.94$  & $2.04$  \\
        DOMINANT   & $81.26$ & $10.16$ & $9.17$  & $88.17$ & $0.67$  & $0.00$  & $83.82$ & $2.13$  & $3.06$  \\
        AnomalyDAE & $74.82$ & $23.90$ & $35.67$ & $72.26$ & $0.84$  & $0.00$  & $73.81$ & $4.02$  & $7.14$  \\
        AHFAN      & $81.24$ & $25.38$ & $48.03$ & $52.26$ & $0.14$  & $0.00$  & $52.34$ & $1.18$  & $11.43$ \\
        SmoothGNN  & $52.63$ & $3.76$  & $5.67$  & $62.97$ & $0.17$  & $0.00$  & $58.00$ & $0.80$  & $1.02$  \\
        HUGE-GAD   & $94.73$ & $55.11$ & $76.67$ & $96.45$ & $7.73$  & $84.21$ & $96.30$ & $26.33$ & $88.78$ \\
        \midrule
        
        & \multicolumn{3}{c}{\cellcolor{green!20} \textbf{Credit ($22.12\%$ Anomaly Ratio)}} & \multicolumn{3}{c}{\cellcolor{blue!20} \textbf{$0.1\%$ Ratio Core}} & \multicolumn{3}{c}{\cellcolor{yellow!20} \textbf{$0.5\%$ Ratio Core}}\\
        MLPAE      & $85.42$ & $19.59$ & $11.87$ & $87.32$ & $0.28$  & $0.00$  & $86.63$ & $1.40$  & $0.00$  \\
        GCNAE      & $91.19$ & $28.33$ & $28.52$ & $93.67$ & $0.66$  & $0.00$  & $93.54$ & $3.13$  & $0.00$  \\
        COLA       & $61.76$ & $6.17$  & $4.63$  & $66.50$ & $0.10$  & $0.00$  & $66.44$ & $0.45$  & $0.42$  \\
        GADNR      & $57.42$ & $7.92$  & $7.46$  & $61.99$ & $0.14$  & $0.00$  & $63.49$ & $0.70$  & $0.84$  \\
        DOMINANT   & $69.71$ & $10.45$ & $12.24$ & $71.91$ & $0.26$  & $2.13$  & $72.71$ & $0.99$  & $1.26$  \\
        AnomalyDAE & OOM     & OOM     & OOM     & OOM     & OOM     & OOM     & OOM     & OOM     & OOM     \\
        AHFAN      & $100.00$& $100.00$& $99.68$ & $100.00$& $100.00$& $100.00$& $100.00$& $100.00$& $100.00$\\
        SmoothGNN  & $79.85$ & $21.08$ & $27.38$ & $77.53$ & $0.30$  & $0.00$  & $75.95$ & $1.67$  & $4.62$  \\
        HUGE-GAD   & $63.01$ & $9.91$  & $12.18$ & $44.32$ & $0.07$  & $0.00$  & $42.82$ & $0.39$  & $0.00$  \\
        \midrule

        & \multicolumn{3}{c}{\cellcolor{green!20} \textbf{Credit ($22.12\%$ Anomaly Ratio)}} & \multicolumn{3}{c}{\cellcolor{blue!20} \textbf{$0.1\%$ Ratio edge}} & \multicolumn{3}{c}{\cellcolor{yellow!20} \textbf{$0.5\%$ Ratio edge}}\\
        MLPAE      & $58.73$ & $25.30$ & $0.71$  & $78.03$ & $0.16$  & $0.00$  & $78.84$ & $0.68$  & $0.00$  \\
        GCNAE      & $66.74$ & $25.26$ & $23.40$ & $89.23$ & $0.15$  & $0.00$  & $91.36$ & $0.60$  & $0.67$  \\
        COLA       & $52.91$ & $23.48$ & $22.78$ & $77.07$ & $0.22$  & $0.00$  & $70.34$ & $0.84$  & $3.33$  \\
        GADNR      & $52.47$ & $23.63$ & $39.12$ & $51.11$ & $0.10$  & $0.00$  & $45.76$ & $0.43$  & $0.00$  \\
        DOMINANT   & $64.30$ & $18.76$ & $15.45$ & $81.71$ & $0.08$  & $0.00$  & $84.61$ & $0.35$  & $0.00$  \\
        AnomalyDAE & $66.83$ & $34.00$ & $37.94$ & $88.87$ & $0.28$  & $0.00$  & $92.07$ & $1.80$  & $0.00$  \\
        AHFAN      & $75.93$ & $46.74$ & $28.93$ & $99.56$ & $24.00$ & $0.00$  & $99.51$ & $36.29$ & $1.79$  \\
        SmoothGNN  & $70.38$ & $38.08$ & $42.90$ & $90.36$ & $0.86$  & $0.00$  & $92.45$ & $3.71$  & $1.33$  \\
        HUGE-GAD   & $52.68$ & $22.22$ & $5.56$  & $49.58$ & $0.09$  & $0.00$  & $48.13$ & $0.43$  & $0.00$  \\
        \bottomrule
    \end{tabular}
    \label{tab:credit_comparison}
\end{table}

\begin{table}[H]
    \centering
    \footnotesize
    \caption{Performance Comparison across Different Anomaly Ratios on Credit. ROC, PRC, and Rec denote AUC-ROC, AUC-PRC, and Recall, respectively.}
    \begin{tabular}{lrrrrrrrrr}
        \toprule
        \textbf{Algorithm} & \textbf{ROC} & \textbf{PRC} & \textbf{Rec} & \textbf{ROC} & \textbf{PRC} & \textbf{Rec} & \textbf{ROC} & \textbf{PRC} & \textbf{Rec} \\
        \midrule
        & \multicolumn{3}{c}{\cellcolor{green!20} \textbf{Credit ($22.12\%$ Anomaly Ratio)}} & \multicolumn{3}{c}{\cellcolor{blue!20} \textbf{$0.1\%$ Ratio random}} & \multicolumn{3}{c}{\cellcolor{yellow!20} \textbf{$0.5\%$ Ratio random}}\\
        MLPAE      & $58.73$ & $25.30$ & $0.71$  & $72.76$ & $0.16$  & $0.00$  & $62.17$ & $0.65$  & $0.00$  \\
        GCNAE      & $66.74$ & $25.26$ & $23.40$ & $76.00$ & $0.17$  & $0.00$  & $63.82$ & $0.62$  & $0.67$  \\
        COLA       & $52.91$ & $23.48$ & $22.78$ & $64.86$ & $0.11$  & $0.00$  & $52.79$ & $0.48$  & $0.67$  \\
        GADNR      & $52.47$ & $23.63$ & $39.12$ & $49.18$ & $0.10$  & $0.00$  & $50.80$ & $0.51$  & $0.00$  \\
        DOMINANT   & $64.30$ & $18.76$ & $15.45$ & $70.22$ & $0.07$  & $0.00$  & $58.44$ & $0.41$  & $0.00$  \\
        AnomalyDAE & $66.83$ & $34.00$ & $37.94$ & $76.27$ & $0.47$  & $3.33$  & $64.62$ & $0.97$  & $2.67$  \\
        AHFAN      & $75.93$ & $46.74$ & $28.93$ & $67.95$ & $0.17$  & $0.00$  & $67.65$ & $0.17$  & $0.00$  \\
        SmoothGNN  & $70.38$ & $38.08$ & $42.90$ & $75.31$ & $0.26$  & $0.00$  & $62.92$ & $0.75$  & $0.67$  \\
        HUGE-GAD   & $52.68$ & $22.22$ & $5.56$  & $49.88$ & $0.09$  & $3.33$  & $47.97$ & $0.45$  & $1.33$  \\
        \bottomrule
    \end{tabular}
    \label{tab:credit_comparison}
\end{table}

\subsection{Results of Incomplete-Data Analysis} \label{sec:incomplete}
\begin{table}[H]
    \centering
    \footnotesize
    \setlength{\tabcolsep}{8pt}
    \caption{Performance Comparison across Different Missing Ratios (Mean Imputation and Median Imputation) on Twitter. ROC, PRC, and Rec denote AUC-ROC, AUC-PRC, and Recall, respectively.}
    \label{tab:twitter_miss_mean_split}   
    \begin{tabular}{lrrrrrrrrr}
        \toprule
        \textbf{Algorithm} & \textbf{ROC} & \textbf{PRC} & \textbf{Rec} & \textbf{ROC} & \textbf{PRC} & \textbf{Rec} & \textbf{ROC} & \textbf{PRC} & \textbf{Rec} \\
        \midrule
        & \multicolumn{3}{c}{\cellcolor{green!20} \textbf{Original Twitter}} & \multicolumn{3}{c}{\cellcolor{blue!20} \textbf{10\% Missing (Mean)}} & \multicolumn{3}{c}{\cellcolor{yellow!20} \textbf{20\% Missing (Mean)}} \\
        MLPAE      & $84.79$ & $19.59$ & $11.87$ & $71.85$ & $12.65$ & $5.57$  & $72.31$ & $13.16$ & $5.76$  \\
        GCNAE      & $91.19$ & $28.33$ & $28.52$ & $91.80$ & $31.40$ & $33.87$ & $92.95$ & $35.18$ & $39.60$ \\
        COLA       & $61.76$ & $6.17$  & $4.63$  & $46.83$ & $6.19$  & $5.00$  & $46.20$ & $6.08$  & $5.35$  \\
        GADNR      & $57.42$ & $7.92$  & $7.46$  & $56.91$ & $7.86$  & $7.46$  & $57.18$ & $7.89$  & $7.46$  \\
        DOMINANT   & $69.71$ & $10.45$ & $12.24$ & $69.01$ & $11.52$ & $13.28$ & $69.53$ & $11.68$ & $13.57$ \\
        AnomalyDAE & OOM     & OOM     & OOM     & OOM     & OOM     & OOM     & OOM     & OOM     & OOM     \\
        AHFAN      & $100.0$ & $100.0$ & $99.68$ & $100.0$ & $100.0$ & $100.0$ & $100.0$ & $100.0$ & $100.0$ \\
        SmoothGNN  & $79.85$ & $21.08$ & $27.38$ & $80.50$ & $22.28$ & $28.67$ & $82.66$ & $25.11$ & $32.07$ \\
        HUGE-GAD   & $63.01$ & $9.91$  & $12.18$ & $67.80$ & $11.92$ & $15.23$ & $73.27$ & $15.32$ & $19.64$ \\
        \midrule
        
        & \multicolumn{3}{c}{\cellcolor{green!20} \textbf{30\% Missing (Mean) }} & \multicolumn{3}{c}{\cellcolor{blue!20} \textbf{40\% Missing (Mean)}} & \multicolumn{3}{c}{\cellcolor{yellow!20} \textbf{50\% Missing (Mean)}} \\
        MLPAE      & $73.67$ & $14.66$ & $6.36$  & $74.24$ & $15.53$ & $7.62$  & $76.42$ & $18.21$ & $8.56$  \\
        GCNAE      & $94.18$ & $39.74$ & $44.79$ & $95.34$ & $45.31$ & $52.12$ & $96.52$ & $52.27$ & $58.51$ \\
        COLA       & $45.91$ & $6.06$  & $5.45$  & $42.21$ & $5.91$  & $6.39$  & $39.18$ & $5.87$  & $6.96$  \\
        GADNR      & $56.98$ & $7.86$  & $7.46$  & $56.71$ & $7.84$  & $7.46$  & $56.83$ & $7.85$  & $7.46$  \\
        DOMINANT   & $70.10$ & $11.87$ & $13.85$ & $70.74$ & $12.08$ & $13.85$ & $71.49$ & $12.34$ & $14.16$ \\
        AnomalyDAE & OOM     & OOM     & OOM     & OOM     & OOM     & OOM     & OOM     & OOM     & OOM     \\
        AHFAN      & $99.99$ & $99.96$ & $99.92$ & $99.92$ & $99.85$ & $99.68$ & $99.93$ & $99.88$ & $99.60$ \\
        SmoothGNN  & $84.90$ & $29.02$ & $35.98$ & $83.23$ & $26.74$ & $33.99$ & $84.24$ & $29.75$ & $37.33$ \\
        HUGE-GAD   & $79.16$ & $21.36$ & $27.29$ & $85.14$ & $31.76$ & $37.27$ & $91.41$ & $51.77$ & $52.88$ \\
        \midrule
        & \multicolumn{3}{c}{\cellcolor{green!20} \textbf{Original Twitter}} & \multicolumn{3}{c}{\cellcolor{blue!20} \textbf{10\% Missing (Median)}} & \multicolumn{3}{c}{\cellcolor{yellow!20} \textbf{20\% Missing (Median)}} \\
        MLPAE      & $84.79$ & $19.59$ & $11.87$ & $71.51$ & $12.49$ & $5.38$  & $72.18$ & $13.33$ & $6.23$  \\
        GCNAE      & $91.19$ & $28.33$ & $28.52$ & $91.98$ & $31.83$ & $34.34$ & $93.23$ & $36.08$ & $40.64$ \\
        COLA       & $61.76$ & $6.17$  & $4.63$  & $46.71$ & $6.17$  & $4.88$  & $46.04$ & $6.06$  & $5.13$  \\
        GADNR      & $57.42$ & $7.92$  & $7.46$  & $57.25$ & $7.89$  & $7.46$  & $57.29$ & $7.90$  & $7.46$  \\
        DOMINANT   & $69.71$ & $10.45$ & $12.24$ & $69.11$ & $11.54$ & $13.35$ & $69.74$ & $11.72$ & $13.63$ \\
        AnomalyDAE & OOM     & OOM     & OOM     & OOM     & OOM     & OOM     & OOM     & OOM     & OOM     \\
        AHFAN      & $100.0$ & $100.0$ & $99.68$ & $100.0$ & $100.0$ & $100.0$ & $100.0$ & $100.0$ & $100.0$ \\
        SmoothGNN  & $79.85$ & $21.08$ & $27.38$ & $79.65$ & $20.99$ & $27.54$ & $80.42$ & $21.76$ & $29.15$ \\
        HUGE-GAD   & $63.01$ & $9.91$  & $12.18$ & $66.18$ & $11.27$ & $14.38$ & $70.45$ & $13.56$ & $17.63$ \\
        \midrule
        
        & \multicolumn{3}{c}{\cellcolor{green!20} \textbf{30\% Missing (Median)}} & \multicolumn{3}{c}{\cellcolor{blue!20} \textbf{40\% Missing (Median)}} & \multicolumn{3}{c}{\cellcolor{yellow!20} \textbf{50\% Missing (Median)}} \\
        MLPAE      & $72.59$ & $14.37$ & $6.11$  & $75.15$ & $16.55$ & $7.59$  & $76.21$ & $18.70$ & $8.03$  \\
        GCNAE      & $94.50$ & $41.23$ & $47.40$ & $95.70$ & $47.43$ & $53.57$ & $96.76$ & $54.60$ & $60.65$ \\
        COLA       & $45.61$ & $5.99$  & $5.63$  & $44.78$ & $6.03$  & $6.86$  & $43.92$ & $6.08$  & $6.89$  \\
        GADNR      & $57.27$ & $7.90$  & $7.46$  & $57.60$ & $7.94$  & $7.46$  & $57.89$ & $7.99$  & $7.46$  \\
        DOMINANT   & $70.42$ & $11.94$ & $13.94$ & $71.20$ & $12.19$ & $14.01$ & $72.05$ & $12.47$ & $14.35$ \\
        AnomalyDAE & OOM     & OOM     & OOM     & OOM     & OOM     & OOM     & OOM     & OOM     & OOM     \\
        AHFAN      & $100.0$ & $100.0$ & $100.0$ & $100.0$ & $99.99$ & $99.84$ & $99.91$ & $99.76$ & $99.36$ \\
        SmoothGNN  & $81.53$ & $22.98$ & $30.88$ & $83.11$ & $24.85$ & $32.55$ & $84.30$ & $26.74$ & $35.10$ \\
        HUGE-GAD   & $76.31$ & $18.44$ & $24.30$ & $81.81$ & $27.20$ & $33.62$ & $87.64$ & $39.72$ & $43.25$ \\
        \bottomrule
    \end{tabular}
\end{table}

\begin{table}[H]
    \centering
    \footnotesize
    \setlength{\tabcolsep}{7pt}
    \caption{Performance Comparison across Different Missing Ratios (Neighbor Imputation) on Twitter and (Mean Imputation and Median Impuataion) on Pubmed. ROC, PRC, and Rec denote AUC-ROC, AUC-PRC, and Recall, respectively.}
    \label{tab:twitter_miss_median_split}
    
    \begin{tabular}{lrrrrrrrrr}
        \toprule
        \textbf{Algorithm} & \textbf{ROC} & \textbf{PRC} & \textbf{Rec} & \textbf{ROC} & \textbf{PRC} & \textbf{Rec} & \textbf{ROC} & \textbf{PRC} & \textbf{Rec} \\
        \midrule     
        & \multicolumn{3}{c}{\cellcolor{green!20} \textbf{Original Twitter}} & \multicolumn{3}{c}{\cellcolor{blue!20} \textbf{10\% Missing (Neighbor)}} & \multicolumn{3}{c}{\cellcolor{yellow!20} \textbf{20\% Missing (Neighbor)}} \\
        MLPAE      & $84.79$ & $19.59$ & $11.87$ & $70.27$ & $12.23$ & $5.73$  & $68.02$ & $11.68$ & $5.57$  \\
        GCNAE      & $91.19$ & $28.33$ & $28.52$ & $91.04$ & $29.72$ & $31.32$ & $91.38$ & $31.38$ & $35.35$ \\
        COLA       & $61.76$ & $6.17$  & $4.63$  & $47.47$ & $6.26$  & $4.97$  & $47.41$ & $6.25$  & $4.97$  \\
        GADNR      & $57.42$ & $7.92$  & $7.46$  & $57.27$ & $7.90$  & $7.46$  & $57.03$ & $7.87$  & $7.46$  \\
        DOMINANT   & $69.71$ & $10.45$ & $12.24$ & $68.47$ & $11.43$ & $13.19$ & $68.46$ & $11.50$ & $13.47$ \\
        AnomalyDAE & OOM     & OOM     & OOM     & OOM     & OOM     & OOM     & OOM     & OOM     & OOM     \\
        AHFAN      & $100.00$& $100.00$& $99.68$ & $100.00$& $100.00$& $100.00$& $100.00$& $99.90$ & $100.00$\\
        SmoothGNN  & $79.85$ & $21.08$ & $27.38$ & $80.06$ & $21.56$ & $27.70$ & $78.81$ & $19.79$ & $26.44$ \\
        HUGE-GAD   & $63.01$ & $9.91$  & $12.18$ & $63.71$ & $10.63$ & $14.10$ & $66.20$ & $12.40$ & $17.06$ \\
        \midrule
        
        & \multicolumn{3}{c}{\cellcolor{green!20} \textbf{30\% Missing (Neighbor)}} & \multicolumn{3}{c}{\cellcolor{blue!20} \textbf{40\% Missing (Neighbor)}} & \multicolumn{3}{c}{\cellcolor{yellow!20} \textbf{50\% Missing (Neighbor)}} \\
        MLPAE      & $69.81$ & $13.21$ & $6.61$  & $68.77$ & $13.37$ & $7.46$  & $68.45$ & $13.96$ & $6.89$  \\
        GCNAE      & $91.82$ & $32.95$ & $37.30$ & $92.47$ & $35.45$ & $40.54$ & $93.38$ & $38.66$ & $44.22$ \\
        COLA       & $47.72$ & $6.33$  & $5.23$  & $47.65$ & $6.27$  & $5.10$  & $48.25$ & $6.31$  & $5.07$  \\
        GADNR      & $56.94$ & $7.86$  & $7.46$  & $57.25$ & $7.89$  & $7.46$  & $57.18$ & $7.88$  & $7.46$  \\
        DOMINANT   & $68.54$ & $11.59$ & $13.63$ & $68.77$ & $11.75$ & $13.85$ & $69.18$ & $11.94$ & $14.23$ \\
        AnomalyDAE & OOM     & OOM     & OOM     & OOM     & OOM     & OOM     & OOM     & OOM     & OOM     \\
        AHFAN      & $100.00$& $99.89$ & $100.00$& $100.00$& $99.87$ & $99.84$ & $99.93$ & $99.72$ & $99.68$ \\
        SmoothGNN  & $79.47$ & $20.95$ & $27.89$ & $81.06$ & $22.45$ & $29.46$ & $83.65$ & $22.33$ & $30.63$ \\
        HUGE-GAD   & $68.58$ & $14.85$ & $19.89$ & $72.14$ & $19.27$ & $24.55$ & $76.33$ & $24.54$ & $29.46$ \\
        \midrule
        
        & \multicolumn{3}{c}{\cellcolor{green!20} \textbf{Original Pubmed}} & \multicolumn{3}{c}{\cellcolor{blue!20} \textbf{10\% Missing (Mean)}} & \multicolumn{3}{c}{\cellcolor{yellow!20} \textbf{20\% Missing (Mean)}} \\
        MLPAE      & $73.07$ & $11.04$ & $6.00$  & $69.54$ & $9.66$  & $6.67$  & $65.61$ & $8.86$  & $6.17$  \\
        GCNAE      & $74.41$ & $23.37$ & $34.50$ & $72.87$ & $20.97$ & $32.83$ & $72.40$ & $20.02$ & $30.83$ \\
        COLA       & $49.81$ & $3.18$  & $3.00$  & $45.63$ & $3.17$  & $2.67$  & $45.83$ & $3.19$  & $2.50$  \\
        GADNR      & $75.75$ & $9.21$  & $8.50$  & $75.39$ & $9.19$  & $8.50$  & $74.55$ & $9.13$  & $8.33$  \\
        DOMINANT   & $81.26$ & $10.16$ & $9.17$  & $80.59$ & $10.04$ & $9.00$  & $80.15$ & $9.98$  & $9.17$  \\
        AnomalyDAE & $74.82$ & $23.90$ & $35.67$ & $73.20$ & $21.27$ & $33.33$ & $72.65$ & $20.20$ & $30.83$ \\
        AHFAN      & $81.24$ & $25.38$ & $48.03$ & $82.97$ & $31.58$ & $60.79$ & $82.66$ & $28.48$ & $52.42$ \\
        SmoothGNN  & $52.63$ & $3.76$  & $5.67$  & $52.62$ & $3.75$  & $6.33$  & $52.76$ & $3.77$  & $6.50$  \\
        HUGE-GAD   & $94.73$ & $55.11$ & $76.67$ & $94.26$ & $52.40$ & $52.33$ & $94.03$ & $52.11$ & $51.83$ \\
        \midrule
        
        & \multicolumn{3}{c}{\cellcolor{green!20} \textbf{30\% Missing (Mean)}} & \multicolumn{3}{c}{\cellcolor{blue!20} \textbf{40\% Missing (Mean)}} & \multicolumn{3}{c}{\cellcolor{yellow!20} \textbf{50\% Missing (Mean)}} \\
        MLPAE      & $62.05$ & $8.34$  & $8.17$  & $57.35$ & $7.54$  & $8.83$  & $59.17$ & $7.50$  & $9.50$  \\
        GCNAE      & $71.48$ & $18.37$ & $28.50$ & $69.06$ & $15.81$ & $25.17$ & $67.15$ & $13.57$ & $22.00$ \\
        COLA       & $49.95$ & $3.73$  & $1.83$  & $48.80$ & $3.61$  & $2.50$  & $46.18$ & $3.46$  & $2.67$  \\
        GADNR      & $74.42$ & $9.12$  & $8.50$  & $74.41$ & $9.14$  & $8.50$  & $73.67$ & $9.09$  & $8.67$  \\
        DOMINANT   & $79.56$ & $9.89$  & $9.00$  & $78.55$ & $9.74$  & $8.67$  & $77.63$ & $9.64$  & $8.50$  \\
        AnomalyDAE & $71.47$ & $18.37$ & $28.67$ & $69.08$ & $15.82$ & $25.00$ & $67.18$ & $13.57$ & $22.00$ \\
        AHFAN      & $85.04$ & $38.82$ & $61.67$ & $89.44$ & $44.52$ & $68.72$ & $91.71$ & $63.07$ & $74.89$ \\
        SmoothGNN  & $53.47$ & $3.94$  & $6.33$  & $52.54$ & $3.96$  & $5.83$  & $51.76$ & $3.72$  & $5.33$  \\
        HUGE-GAD   & $93.70$ & $49.25$ & $49.67$ & $92.81$ & $44.68$ & $46.00$ & $91.74$ & $40.51$ & $42.00$ \\
        \midrule
        & \multicolumn{3}{c}{\cellcolor{green!20} \textbf{Original Pubmed}} & \multicolumn{3}{c}{\cellcolor{blue!20} \textbf{10\% Missing (Median)}} & \multicolumn{3}{c}{\cellcolor{yellow!20} \textbf{20\% Missing (Median)}} \\
        MLPAE      & 73.07 & 11.04 & 6.00  & 70.59 & 9.72  & 6.50  & 69.61 & 9.42  & 6.17  \\
        GCNAE      & 74.41 & 23.37 & 34.50 & 72.76 & 20.92 & 32.83 & 72.13 & 19.92 & 30.67 \\
        COLA       & 49.81 & 3.18  & 3.00  & 47.10 & 3.50  & 1.83  & 50.31 & 3.43  & 3.50  \\
        GADNR      & 75.75 & 9.21  & 8.50  & 75.76 & 9.12  & 8.50  & 75.62 & 9.21  & 8.50  \\
        DOMINANT   & 81.26 & 10.16 & 9.17  & 80.57 & 10.03 & 9.00  & 80.12 & 9.96  & 9.17  \\
        AnomalyDAE & 74.82 & 23.90 & 35.67 & 73.11 & 21.29 & 33.50 & 72.42 & 20.17 & 30.50 \\
        AHFAN      & 81.24 & 25.38 & 48.03 & 81.81 & 27.72 & 51.98 & 80.19 & 24.93 & 49.78 \\
        SmoothGNN  & 52.63 & 3.76  & 5.67  & 52.41 & 3.72  & 6.00  & 52.25 & 3.70  & 5.67  \\
        HUGE-GAD   & 94.73 & 55.11 & 76.67 & 94.26 & 53.00 & 53.17 & 93.96 & 52.83 & 53.17 \\
        \midrule
        & \multicolumn{3}{c}{\cellcolor{green!20} \textbf{30\% Missing (Median)}} & \multicolumn{3}{c}{\cellcolor{blue!20} \textbf{40\% Missing (Median)}} & \multicolumn{3}{c}{\cellcolor{yellow!20} \textbf{50\% Missing (Median)}} \\
        MLPAE      & 68.06 & 9.08  & 7.67  & 65.90 & 8.61  & 9.00  & 64.47 & 8.35  & 8.50  \\
        GCNAE      & 71.00 & 18.22 & 28.33 & 68.30 & 15.62 & 24.83 & 65.93 & 13.32 & 21.83 \\
        COLA       & 50.25 & 3.68  & 3.00  & 49.50 & 3.69  & 2.00  & 50.73 & 3.80  & 2.17  \\
        GADNR      & 75.55 & 9.16  & 8.50  & 75.46 & 9.17  & 8.50  & 75.09 & 9.18  & 8.50  \\
        DOMINANT   & 79.49 & 9.87  & 8.83  & 78.44 & 9.71  & 8.67  & 77.46 & 9.60  & 8.50  \\
        AnomalyDAE & 71.19 & 18.40 & 28.33 & 68.19 & 15.59 & 25.00 & 65.78 & 13.29 & 21.67 \\
        AHFAN      & 83.08 & 35.83 & 65.20 & 79.97 & 28.89 & 50.66 & 81.04 & 31.39 & 48.46 \\
        SmoothGNN  & 52.60 & 3.85  & 5.67  & 52.04 & 3.95  & 5.67  & 51.35 & 3.73  & 4.67  \\
        HUGE-GAD   & 93.41 & 50.13 & 50.50 & 92.49 & 45.36 & 47.00 & 91.11 & 41.24 & 42.00 \\
        \bottomrule
    \end{tabular}
\end{table}

\begin{table}[H]
    \centering
    \footnotesize
    \setlength{\tabcolsep}{9pt}
    \caption{Performance Comparison across Different Missing Ratios (Neighbor Imputation) on Pubmed and (Mean Imputation and Median Imputation) on Credit. ROC, PRC, and Rec denote AUC-ROC, AUC-PRC, and Recall, respectively.}
    \begin{tabular}{lrrrrrrrrr}
        \toprule
        \textbf{Algorithm} & \textbf{ROC} & \textbf{PRC} & \textbf{Rec} & \textbf{ROC} & \textbf{PRC} & \textbf{Rec} & \textbf{ROC} & \textbf{PRC} & \textbf{Rec} \\
        \midrule
        & \multicolumn{3}{c}{\cellcolor{green!20} \textbf{Original Pubmed}} & \multicolumn{3}{c}{\cellcolor{blue!20} \textbf{10\% Missing (Neighbor)}} & \multicolumn{3}{c}{\cellcolor{yellow!20} \textbf{20\% Missing (Neighbor)}} \\
        MLPAE      & 73.07 & 11.04 & 6.00  & 68.36 & 8.78  & 4.67  & 64.52 & 7.28  & 4.17  \\
        GCNAE      & 74.41 & 23.37 & 34.50 & 70.75 & 19.37 & 31.00 & 67.95 & 16.77 & 27.33 \\
        COLA       & 49.81 & 3.18  & 3.00  & 48.42 & 3.48  & 1.83  & 49.17 & 3.45  & 2.17  \\
        GADNR      & 75.75 & 9.21  & 8.50  & 75.54 & 9.20  & 8.50  & 75.26 & 9.18  & 8.50  \\
        DOMINANT   & 81.26 & 10.16 & 9.17  & 80.31 & 10.00 & 9.00  & 79.53 & 9.90  & 9.17  \\
        AnomalyDAE & 74.82 & 23.90 & 35.67 & 71.26 & 19.89 & 32.33 & 68.48 & 17.16 & 27.50 \\
        AHFAN      & 81.24 & 25.38 & 48.03 & 81.53 & 26.86 & 52.86 & 82.19 & 28.91 & 51.98 \\
        SmoothGNN  & 52.63 & 3.76  & 5.67  & 52.98 & 3.63  & 5.33  & 53.55 & 3.57  & 5.00  \\
        HUGE-GAD   & 94.73 & 55.11 & 76.67 & 95.35 & 58.96 & 56.50 & 96.09 & 64.49 & 61.50 \\
        \midrule
        & \multicolumn{3}{c}{\cellcolor{green!20} \textbf{30\% Missing (Neighbor)}} & \multicolumn{3}{c}{\cellcolor{blue!20} \textbf{40\% Missing (Neighbor)}} & \multicolumn{3}{c}{\cellcolor{yellow!20} \textbf{50\% Missing (Neighbor)}} \\
        MLPAE      & 60.08 & 6.07  & 3.83  & 54.00 & 4.93  & 4.33  & 48.39 & 4.08  & 4.33  \\
        GCNAE      & 64.71 & 14.24 & 23.83 & 59.82 & 11.50 & 21.17 & 55.57 & 9.20  & 17.33 \\
        COLA       & 51.19 & 3.58  & 2.17  & 52.52 & 3.43  & 2.17  & 50.56 & 3.17  & 1.67  \\
        GADNR      & 75.04 & 9.16  & 8.50  & 75.15 & 9.16  & 8.50  & 74.43 & 9.00  & 8.33  \\
        DOMINANT   & 78.55 & 9.79  & 8.83  & 77.14 & 9.63  & 8.67  & 75.68 & 9.51  & 8.50  \\
        AnomalyDAE & 65.27 & 14.58 & 24.17 & 60.50 & 11.83 & 20.83 & 56.38 & 9.43  & 18.17 \\
        AHFAN      & 82.45 & 34.58 & 61.67 & 84.48 & 45.19 & 60.35 & 87.51 & 45.22 & 63.00 \\
        SmoothGNN  & 54.11 & 3.58  & 4.33  & 53.68 & 3.56  & 4.00  & 54.28 & 3.52  & 3.17  \\
        HUGE-GAD   & 96.62 & 68.02 & 63.17 & 96.73 & 70.59 & 66.00 & 96.87 & 66.66 & 61.83 \\
        \midrule
        & \multicolumn{3}{c}{\cellcolor{green!20} \textbf{Original Credit}} & \multicolumn{3}{c}{\cellcolor{blue!20} \textbf{10\% Missing (Mean)}} & \multicolumn{3}{c}{\cellcolor{yellow!20} \textbf{20\% Missing (Mean)}} \\
        MLPAE      & 58.73 & 25.30 & 0.71  & 56.58 & 25.89 & 0.71  & 57.56 & 26.32 & 1.12  \\
        GCNAE      & 66.74 & 25.26 & 23.40 & 54.16 & 25.36 & 23.84 & 54.95 & 25.79 & 24.68 \\
        COLA       & 52.91 & 23.48 & 22.78 & 51.97 & 23.11 & 22.26 & 51.12 & 22.68 & 21.05 \\
        GADNR      & 52.47 & 23.63 & 39.12 & 48.63 & 21.33 & 40.31 & 49.38 & 21.57 & 33.14 \\
        DOMINANT   & 64.30 & 18.76 & 15.45 & 42.53 & 18.62 & 15.28 & 42.55 & 18.65 & 15.31 \\
        AnomalyDAE & 66.83 & 34.00 & 37.94 & 66.29 & 33.76 & 40.79 & 67.41 & 34.74 & 41.40 \\
        AHFAN      & 75.93 & 46.74 & 28.93 & 76.82 & 49.76 & 41.01 & 81.60 & 60.70 & 58.62 \\
        SmoothGNN  & 70.38 & 38.08 & 42.90 & 71.15 & 39.24 & 44.41 & 71.79 & 40.50 & 45.74 \\
        HUGE-GAD   & 52.68 & 22.22 & 5.56  & 48.40 & 20.63 & 18.64 & 45.79 & 19.94 & 18.38 \\
        \midrule
        & \multicolumn{3}{c}{\cellcolor{green!20} \textbf{30\% Missing (Mean)}} & \multicolumn{3}{c}{\cellcolor{blue!20} \textbf{40\% Missing (Mean)}} & \multicolumn{3}{c}{\cellcolor{yellow!20} \textbf{50\% Missing (Mean)}} \\
        MLPAE      & 58.52 & 26.72 & 1.33  & 59.19 & 27.16 & 2.03  & 59.80 & 27.69 & 3.56  \\
        GCNAE      & 55.98 & 26.39 & 26.33 & 57.37 & 27.26 & 27.77 & 58.88 & 28.26 & 29.45 \\
        COLA       & 49.96 & 22.23 & 21.26 & 49.11 & 21.88 & 21.17 & 47.86 & 21.59 & 20.09 \\
        GADNR      & 52.32 & 23.38 & 29.08 & 50.49 & 22.05 & 29.46 & 54.33 & 24.04 & 43.16 \\
        DOMINANT   & 42.60 & 18.68 & 15.22 & 42.69 & 18.72 & 15.19 & 42.81 & 18.78 & 15.30 \\
        AnomalyDAE & 69.31 & 36.93 & 43.35 & 70.73 & 37.99 & 43.35 & 72.02 & 39.30 & 45.54 \\
        AHFAN      & 84.61 & 71.84 & 57.83 & 91.48 & 84.63 & 72.54 & 99.09 & 97.37 & 91.76 \\
        SmoothGNN  & 72.20 & 41.58 & 46.79 & 72.87 & 42.87 & 47.36 & 73.26 & 44.04 & 47.98 \\
        HUGE-GAD   & 44.91 & 20.52 & 17.07 & 44.93 & 21.04 & 18.17 & 44.58 & 21.69 & 20.00 \\
        \midrule
        & \multicolumn{3}{c}{\cellcolor{green!20} \textbf{Original Credit}} & \multicolumn{3}{c}{\cellcolor{blue!20} \textbf{10\% Missing (Median)}} & \multicolumn{3}{c}{\cellcolor{yellow!20} \textbf{20\% Missing (Median)}} \\
        MLPAE      & 58.73 & 25.30 & 0.71  & 58.08 & 27.13 & 2.11  & 60.49 & 28.82 & 2.95  \\
        GCNAE      & 66.74 & 25.26 & 23.40 & 55.73 & 26.55 & 25.90 & 58.06 & 28.59 & 29.75 \\
        COLA       & 52.91 & 23.48 & 22.78 & 52.33 & 23.35 & 22.42 & 52.14 & 23.34 & 21.99 \\
        GADNR      & 52.47 & 23.63 & 39.12 & 50.37 & 22.28 & 39.80 & 48.49 & 22.48 & 40.61 \\
        DOMINANT   & 64.30 & 18.76 & 15.45 & 42.69 & 18.79 & 15.40 & 42.85 & 18.99 & 15.27 \\
        AnomalyDAE & 66.83 & 34.00 & 37.94 & 65.43 & 30.73 & 23.15 & 73.57 & 42.07 & 48.76 \\
        AHFAN      & 75.93 & 46.74 & 28.93 & 77.75 & 53.00 & 55.57 & 81.92 & 60.80 & 57.41 \\
        SmoothGNN  & 70.38 & 38.08 & 42.90 & 72.13 & 40.63 & 45.87 & 74.04 & 43.85 & 48.48 \\
        HUGE-GAD   & 52.68 & 22.22 & 5.56  & 51.92 & 23.18 & 19.61 & 55.59 & 26.89 & 23.88 \\
        \midrule
        & \multicolumn{3}{c}{\cellcolor{green!20} \textbf{30\% Missing (Median)}} & \multicolumn{3}{c}{\cellcolor{blue!20} \textbf{40\% Missing (Median)}} & \multicolumn{3}{c}{\cellcolor{yellow!20} \textbf{50\% Missing (Median)}} \\
        MLPAE      & 62.73 & 30.34 & 2.86  & 64.98 & 32.26 & 3.19  & 66.41 & 33.76 & 4.45  \\
        GCNAE      & 60.83 & 31.43 & 32.73 & 64.28 & 35.46 & 37.16 & 68.51 & 40.95 & 41.67 \\
        COLA       & 51.54 & 23.16 & 21.53 & 51.37 & 23.19 & 21.75 & 50.21 & 22.76 & 20.60 \\
        GADNR      & 48.84 & 22.57 & 36.17 & 46.22 & 21.00 & 40.10 & 46.34 & 20.89 & 34.16 \\
        DOMINANT   & 43.10 & 19.19 & 15.39 & 43.47 & 19.45 & 15.39 & 44.02 & 19.82 & 15.48 \\
        AnomalyDAE & 78.26 & 49.30 & 53.71 & 82.53 & 56.13 & 58.56 & 86.65 & 64.00 & 65.28 \\
        AHFAN      & 94.56 & 86.56 & 78.97 & 97.19 & 92.26 & 80.59 & 98.66 & 95.62 & 92.36 \\
        SmoothGNN  & 76.02 & 47.01 & 51.10 & 78.57 & 50.13 & 53.68 & 81.52 & 53.92 & 57.08 \\
        HUGE-GAD   & 58.44 & 28.65 & 25.92 & 60.45 & 30.68 & 27.83 & 62.02 & 33.89 & 30.92 \\
        \bottomrule
    \end{tabular}
\end{table}

\begin{table}[H]
    \centering
    \footnotesize
    \setlength{\tabcolsep}{9pt}
    \caption{Performance Comparison across Different Missing Ratios (Neighbor Imputation) on Credit. ROC, PRC, and Rec denote AUC-ROC, AUC-PRC, and Recall, respectively.}
    \begin{tabular}{lrrrrrrrrr}
        \toprule
        \textbf{Algorithm} & \textbf{ROC} & \textbf{PRC} & \textbf{Rec} & \textbf{ROC} & \textbf{PRC} & \textbf{Rec} & \textbf{ROC} & \textbf{PRC} & \textbf{Rec} \\
        \midrule
        & \multicolumn{3}{c}{\cellcolor{green!20} \textbf{Original Credit}} & \multicolumn{3}{c}{\cellcolor{blue!20} \textbf{10\% (Neighbor)}} & \multicolumn{3}{c}{\cellcolor{yellow!20} \textbf{20\% (Neighbor)}} \\
        MLPAE      & 58.73 & 25.30 & 0.71  & 55.85 & 25.64 & 0.81  & 56.31 & 26.03 & 0.93  \\
        GCNAE      & 66.74 & 25.26 & 23.40 & 54.54 & 25.65 & 23.88 & 55.27 & 26.17 & 24.47 \\
        COLA       & 52.91 & 23.48 & 22.78 & 52.67 & 23.51 & 22.80 & 52.75 & 23.68 & 23.03 \\
        GADNR      & 52.47 & 23.63 & 39.12 & 49.99 & 21.91 & 37.51 & 48.56 & 21.09 & 34.86 \\
        DOMINANT   & 64.30 & 18.76 & 15.45 & 42.63 & 18.64 & 15.33 & 42.75 & 18.67 & 15.30 \\
        AnomalyDAE & 66.83 & 34.00 & 37.94 & 65.81 & 33.89 & 40.46 & 66.30 & 34.44 & 40.82 \\
        AHFAN      & 75.93 & 46.74 & 28.93 & 76.06 & 47.00 & 29.83 & 76.19 & 47.24 & 29.35 \\
        SmoothGNN  & 70.38 & 38.08 & 42.90 & 70.46 & 38.21 & 43.01 & 70.53 & 38.36 & 43.22 \\
        HUGE-GAD   & 52.68 & 22.22 & 5.56  & 52.91 & 22.44 & 19.15 & 53.09 & 22.76 & 19.83 \\
        \midrule
        & \multicolumn{3}{c}{\cellcolor{green!20} \textbf{30\% Missing (Neighbor)}} & \multicolumn{3}{c}{\cellcolor{blue!20} \textbf{40\% Missing (Neighbor)}} & \multicolumn{3}{c}{\cellcolor{yellow!20} \textbf{50\% Missing (Neighbor)}} \\
        MLPAE      & 56.81 & 26.37 & 1.28  & 57.17 & 26.63 & 1.64  & 57.38 & 26.90 & 2.17  \\
        GCNAE      & 56.14 & 26.72 & 25.00 & 57.25 & 27.59 & 25.98 & 58.51 & 28.47 & 26.72 \\
        COLA       & 52.70 & 23.77 & 23.42 & 52.77 & 23.96 & 23.70 & 52.82 & 24.13 & 23.85 \\
        GADNR      & 51.95 & 23.28 & 42.72 & 51.72 & 22.88 & 37.39 & 53.11 & 23.39 & 47.69 \\
        DOMINANT   & 42.86 & 18.72 & 15.39 & 43.03 & 18.76 & 15.45 & 43.24 & 18.83 & 15.40 \\
        AnomalyDAE & 66.87 & 35.17 & 43.04 & 67.32 & 35.72 & 43.81 & 67.65 & 36.14 & 44.51 \\
        AHFAN      & 76.96 & 48.88 & 34.42 & 77.28 & 49.43 & 36.98 & 78.49 & 51.26 & 40.71 \\
        SmoothGNN  & 70.60 & 38.45 & 43.25 & 70.62 & 38.42 & 43.40 & 70.64 & 38.55 & 43.51 \\
        HUGE-GAD   & 53.29 & 23.07 & 20.34 & 53.96 & 23.80 & 20.87 & 54.52 & 24.65 & 21.61 \\
        \bottomrule
    \end{tabular}
\end{table}


\end{document}